\documentclass[11pt]{article}

\usepackage[final]{acl}

\usepackage{times}
\usepackage{latexsym}
\usepackage[T1]{fontenc}
\usepackage[utf8]{inputenc}
\usepackage{microtype}
\usepackage{inconsolata}
\usepackage{graphicx}
\usepackage{booktabs}
\usepackage{multirow}
\usepackage{amsmath,amssymb}
\usepackage{array}
\usepackage{xurl}
\usepackage{xspace}
\usepackage{tikz}
\usetikzlibrary{arrows.meta,calc}

\newcommand{\w}{\textsc{Width}\xspace}
\newcommand{\dpth}{\textsc{Depth}\xspace}
\newcommand{\wlone}{W-L1\xspace}
\newcommand{\wltwo}{W-L2\xspace}
\newcommand{\wlthree}{W-L3\xspace}
\newcommand{\dlone}{D-L1\xspace}
\newcommand{\dltwo}{D-L2\xspace}
\newcommand{\dlthree}{D-L3\xspace}

\title{Disentangling Topology and Diversity in Multi-Agent LLMs for Multilingual Low-Resource Emotion Detection}
\author{
\textbf{Ulugbek Shernazarov$^{1}$} \quad
\textbf{Charitha Ruwansiri Weerakon Basnayake$^{2}$} \\
\textbf{Abdelkhaleq El Jarjini$^{1}$} \quad
\textbf{Noel Crespi$^{1}$} \quad
\textbf{Praboda Rajapaksha$^{1,3}$} \\
$^{1}$Samovar, Telecom SudParis, Institut Polytechnique de Paris, 91120 Palaiseau, France \\
$^{2}$Cardiff Metropolitan University, Cardiff, United Kingdom \\
$^{3}$Deparment of Computer Science, Aberystwyth University, Aberystwyth, United Kingdom \\
\texttt{ulugbek.shernazarov@telecom-sudparis.eu}
}

\hypersetup{pdftitle={Disentangling Topology and Diversity in Multi-Agent LLMs for Multilingual Low-Resource Emotion Detection},pdfauthor={Ulugbek Shernazarov; Charitha Ruwansiri Weerakon Basnayake; Abdelkhaleq El Jarjini; Noel Crespi; Praboda Rajapaksha}}
\begin{document}
\maketitle

\begin{abstract}
Multi-agent LLM systems combine multiple inference calls, but prior work often confounds how calls are connected with how they are diversified. We study these factors independently: inference topology and source of inter-agent diversity. In a controlled \(2 \times 3\) matrix, we cross parallel aggregation and sequential refinement with stochastic sampling, role prompting, and learned QLoRA specialization, under a fixed three-call budget and output protocol within each backbone. Using Qwen2.5-14B-Instruct and Llama-3.1-8B-Instruct, we evaluate all six configurations on multilingual low-resource emotion detection across nine languages. Parallel learned specialization is strongest on Qwen at 52.83 Macro-F1 and reaches 52.94 on Llama. On Qwen it also exceeds same-backbone zero-shot, few-shot, CoT, and seven-call self-consistency baselines. The preferred topology depends on diversity source: sequential refinement helps stochastic and prompted settings, while the learned Width advantage shrinks from 2.83 points on Qwen to 0.17 on Llama. Depth-wise analysis suggests that later learned specialists can overwrite correct early predictions, although the aggregate effect is backbone-dependent. Overall, how agents are differentiated produces larger performance shifts than topology, which should be evaluated jointly with specialization.
\end{abstract}

\section{Introduction}

Multi-agent LLM systems improve prediction quality by allocating multiple inference calls to the same input. A system may sample several independent answers and vote, assign different roles to parallel agents, or pass an answer through a sequence of critique and calibration steps \citep{wang2023selfconsistency,du2024debate,madaan2023selfrefine,shinn2023reflexion,wang2025mixture}. These systems are often grouped under the same ``multi-agent'' label, but they encode different design choices about why multiple calls help. One choice concerns \emph{topology}: should calls be connected in parallel, so that independent errors can cancel under aggregation, or in sequence, so that later calls can revise earlier reasoning? A second choice concerns \emph{diversity}: do agents differ through stochastic decoding, prompt variation, or trained parameters?

This conflation makes reported gains difficult to interpret. Parallel systems often combine aggregation with sampling variation, role prompting, or heterogeneous models, while sequential systems typically vary both feedback structure and role design. As a result, it is often unclear whether improvements arise from the way calls are connected, from the source of inter-agent diversity, or simply from the strength of the underlying model. The distinction matters especially under a fixed inference budget, where one must decide whether additional calls are better spent on voting, revision, prompt engineering or learned specialization.

\begin{figure*}[t]
\centering
\resizebox{0.95\textwidth}{!}{
\begin{tikzpicture}[
    font=\rmfamily\scriptsize,
    >={Latex[length=1.9mm,width=1.25mm]},
    agent/.style={draw=black,line width=0.6pt,minimum width=1.40cm,
                  minimum height=0.65cm,inner sep=1pt,align=center},
    inputbox/.style={draw=black,line width=0.6pt,minimum width=1.90cm,
                  minimum height=0.90cm,inner sep=2pt,align=center},
    mechbox/.style={draw=black,line width=0.6pt,minimum width=5.20cm,
                  inner sep=5pt,align=center},
    spine/.style={draw=black,line width=0.7pt},
    solidflow/.style={->,solid,line width=0.7pt},
    dashflow/.style={->,dashed,dash pattern=on 2.6pt off 1.8pt,line width=0.7pt},
    lab/.style={fill=white,inner sep=1.2pt},
    note/.style={font=\rmfamily\scriptsize},
    strip/.style={font=\rmfamily\fontsize{7}{8.4}\selectfont}
]

% ---------------- title ----------------
\node[font=\rmfamily\bfseries\large] at (7.97,8.35) {Topology $\times$ Diversity Setting};

% ---------------- input ----------------
\node[inputbox] (input) at (1.05,4.75)
    {\textbf{Input}\\[1.5pt] $u_i=(x_i,L_i,b_{L_i})$};
\coordinate (fork) at (2.35,4.75);
\draw[spine] (input.east) -- (fork);

% ---------------- width ----------------
\node[font=\rmfamily\bfseries\small] at (8.10,7.62) {WIDTH (parallel)};
\node[agent] (w1) at (4.35,6.75) {$A_1$};
\node[agent] (w2) at (4.35,5.85) {$A_2$};
\node[agent] (w3) at (4.35,4.95) {$A_3$};

\draw[spine] (2.35,4.75) -- (2.35,6.75);
\draw[solidflow] (2.35,6.75) -- (w1.west);
\draw[solidflow] (2.35,5.85) -- (w2.west);
\draw[solidflow] (2.35,4.95) -- (w3.west);

\node[mechbox,minimum height=2.40cm,anchor=west] (vote) at (10.60,5.85)
{$\hat y_{i,e}^{W}=\mathbf{1}\!\left[\displaystyle\sum_{k=1}^{3}
  \hat y_{i,e}^{(k)}\ge 2\right]$\\[5pt]
 $e\in\mathcal{E},\qquad |\mathcal{E}|=6$};

\draw[dashflow] (w1.east) -- node[lab,above] {$\hat y_i^{(1)}$} (vote.west |- w1.east);
\draw[dashflow] (w2.east) -- node[lab,above] {$\hat y_i^{(2)}$} (vote.west |- w2.east);
\draw[dashflow] (w3.east) -- node[lab,above] {$\hat y_i^{(3)}$} (vote.west |- w3.east);

\node[note] at ($(vote.south)+(0,-0.32)$) {$T=0.3$ when $\delta=\mathrm{L3}$};

% ---------------- depth ----------------
\node[font=\rmfamily\bfseries\small] at (8.10,3.55) {DEPTH (sequential)};
\node[agent] (d1) at (4.35,2.45) {$A_1$};
\node[agent] (d2) at (6.60,2.45) {$A_2$};
\node[agent] (d3) at (8.80,2.45) {$A_3$};

\draw[spine] (2.35,4.75) -- (2.35,2.45);
\draw[solidflow] (2.35,2.45) -- (d1.west);
\draw[dashflow] (d1.east) -- node[lab,above] {$z_i^{(1)}$} (d2.west);
\draw[dashflow] (d2.east) -- node[lab,above] {$z_i^{(2)}$} (d3.west);

\node[mechbox,minimum height=1.90cm,anchor=west] (refine) at (10.60,2.45)
{$z_i^{(l)}=f^{\delta}_{l}\!\left(u_i,z_i^{(<l)};T_l\right),\quad l=1,2,3$\\[4pt]
 $z_i^{(<1)}=\emptyset,\qquad \hat y_i^{D}=\pi\!\left(z_i^{(3)}\right)$};

\draw[dashflow] (d3.east) -- node[lab,above] {$z_i^{(3)}$} (refine.west);

\node[note] at ($(refine.south)+(0,-0.32)$) {$T_1=0.7>T_2=0.5>T_3=0.3$};

% ---------------- key ----------------
\draw[solidflow] (0.20,1.78) -- (0.70,1.78);
\node[strip,anchor=west] at (0.80,1.78) {input flow};
\draw[dashflow] (0.20,1.32) -- (0.70,1.32);
\node[strip,anchor=west] at (0.80,1.32) {model output};

% ---------------- bottom strip ----------------
\draw[line width=0.6pt] (0.10,0.86) -- (15.85,0.86);
\node[strip,anchor=west] at (0.10,0.58)
{\textbf{Diversity source} $\delta\in\{\mathrm{L1},\mathrm{L2},\mathrm{L3}\}$\quad
 $\mathrm{L1}$: seed $s_k$\quad
 $\mathrm{L2}$: prompt $p_k$\quad
 $\mathrm{L3}$: QLoRA adapter $a_k$};
\node[strip,anchor=west] at (0.10,0.20)
{Fixed across all six cells: backbone, three calls per example,
 output schema, reasoning protocol.};

\end{tikzpicture}
}
\caption{Topology $\times$ diversity framework. Width issues three parallel
predictions and combines them by per-label majority vote; Depth issues three
sequential refinement steps and reads out the final state. The diversity source
$\delta$ varies across stochastic, prompted, and learned conditions, while the
backbone, the three-call budget, the output schema, and the reasoning protocol
are held fixed in every cell.}
\label{fig:topology-diversity}
\end{figure*}
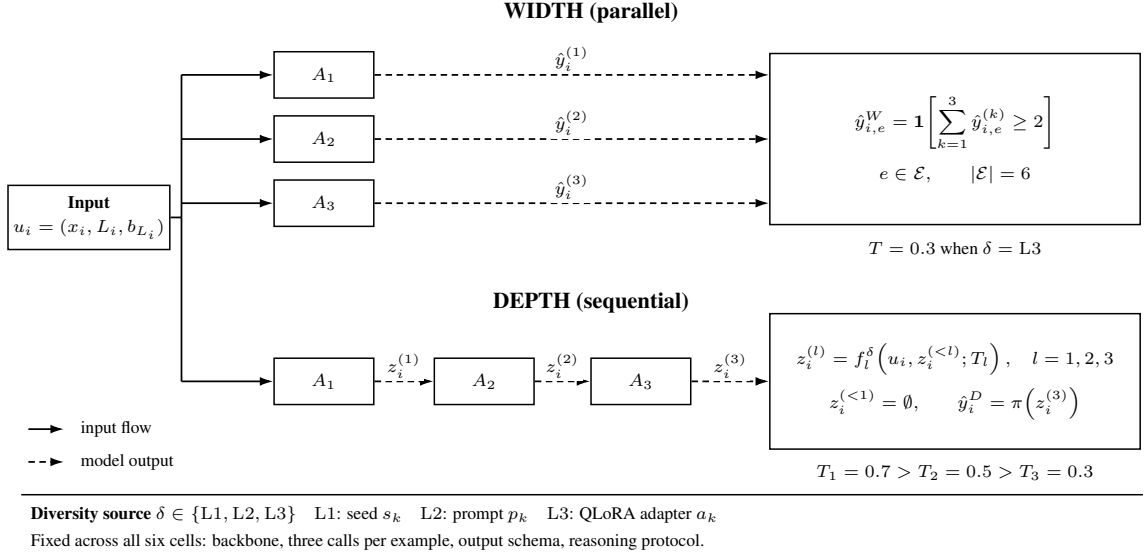

Fine-grained emotion detection is a useful setting for this comparison because the task separates evidence gathering from decision aggregation. Emotional meaning may depend on lexical cues, emojis, sarcasm, negation, pragmatic context, and multiple co-occurring emotions. These cases make topology and diversity testable: different calls can attend to different evidence while the final output remains a fixed multi-label decision. Recent multi-agent work on emotion analysis instantiated a single coordination architecture rather than isolating which design choice is responsible for the gain \citep{dong2026emotion}. We study this question through a controlled \(2 \times 3\) matrix, (Figure~\ref{fig:topology-diversity} and Section~\ref{sec:method}). The topology axis compares \emph{Width} - three agents predict independently and are aggregated in parallel, with \emph{Depth} - three calls form a sequential refinement chain. 
We instantiate the complete six-cell matrix independently on Qwen2.5-14B-Instruct \citep{yang2024qwen25} and Llama-3.1-8B-Instruct \citep{grattafiori2024llama3}. Within each backbone, we hold the model, call budget, output schema, and reasoning protocol fixed, allowing topology and diversity source to be studied as separate experimental factors rather than entangled components of a single architecture.

% ---------------------------------------------------------------
% Figure 1 -- Topology x Diversity framework
% Preamble requirements:
%   \usepackage{tikz}
%   \usetikzlibrary{arrows.meta,calc}
%   \usepackage{amsmath}
%
% Measured width: 452.3pt (15.95cm). ACL \textwidth is 455.2pt (16cm),
% so this fits a figure* with ~3pt to spare. If your style file uses a
% narrower text block, wrap the tikzpicture in
%   \resizebox{\textwidth}{!}{ ... }
% ---------------------------------------------------------------

We evaluate this design on multilingual low-resource emotion detection \citep{muhammad2025brighter} across nine languages. The setting also exposes a failure mode that English-only evaluation can obscure: when the base model has weak language competence, parallel voting among weak agents may reproduce shared errors or abstention rather than create useful diversity. The same fixed six-label task is evaluated across languages with widely different base-model competence, creating a natural stress test for whether additional calls contribute complementary evidence or merely repeat shared cross-lingual failures.
Contributions in this paper are within a two-backbone, fixed three-call setting. First, it provides a controlled comparison of topology and diversity source in multi-agent LLM inference under fixed compute within each backbone. Second, it shows across Qwen2.5-14B and Llama-3.1-8B that learned specialization via lightweight adapters is substantially more effective than stochastic or prompt-based diversification, while explicitly separating the effect of pooled supervised adaptation from specialist heterogeneity. Third, it gives a mechanistic account of when sequential refinement helps and when it hurts: depth is useful when later calls can add evidence to weak initial predictions, while the learned Width--Depth ordering is model-dependent and later correction-trained specialists require stronger preservation-aware calibration once the first specialist is already strong.

\section{Related Work}
\label{sec:related}

\paragraph{Multi-agent LLM inference.}
Multi-agent LLM systems improve predictions by making multiple calls to the same or related inputs, but prior work often changes the connection pattern and the source of diversity at the same time. Chain-of-thought prompting elicits intermediate reasoning \citep{wei2022cot,kojima2022zero}; self-consistency samples several reasoning traces and aggregates final answers \citep{wang2023selfconsistency}, while debate and mixture-of-agents systems combine critique, aggregation, proposer identity, or heterogeneous models \citep{du2024debate,wang2025mixture}. Sequential methods such as Self-Refine and Reflexion instead revise a single trajectory through feedback or reflection \citep{madaan2023selfrefine,shinn2023reflexion}. These systems show that multiple calls can help, but they do not isolate whether gains come from topology, from diversity among calls, or from model strength. This distinction is important because later calls are not automatically beneficial: self-correction can degrade performance when feedback is weak or uninformative \citep{huang2024selfcorrect}. Our work addresses this confound by crossing topology and diversity source while holding the backbone, call budget, output schema, and evaluation setting fixed.

\paragraph{Sources of diversity.}
Diversity in multi-call inference can be induced by stochastic decoding, by prompts or roles, or by trained parameters. Stochastic diversity underlies self-consistency; prompt and role diversity are common in persona-based and collaborative prompting \citep{kong2024roleplay,salewski2023impersonation,kim2025persona}; and parameter diversity can be introduced efficiently through LoRA/QLoRA adapters \citep{hu2022lora,dettmers2023qlora}. Ensemble theory suggests that aggregation is useful only when members are both individually competent and meaningfully diverse \citep{krogh1994ensembles,wood2023diversity}. This motivates our comparison of stochastic, designed and learned diversity under the same three-call budget. In particular, the L3 condition combines parameter diversity with pooled supervised specialization and may therefore improve individual competence as well as inter-agent heterogeneity, whereas prompt-only diversity may mainly vary outputs around the same shared errors. We use the term \emph{learned specialization} when attributing the source of the L3 gain.

\paragraph{Multilingual low-resource emotion detection.}
Fine-grained emotion detection is a natural testbed for this question because emotional meaning can depend on lexical cues, pragmatic context, sarcasm, script, and culturally specific affective expressions. Multilingual emotion benchmarks such as BRIGHTER extend earlier African and multilingual sentiment resources to multi-label emotion classification \citep{muhammad2022naijasenti,muhammad2023afrisenti}. Performance in this setting is shaped by typology, script overlap, pretraining coverage, and resource level \citep{pires2019bert,conneau2020xlmr,wu2020are,lin2019langrank,joshi2020state}. Recent work further suggests that multilingual LLM reasoning may transfer unevenly and may route through English even for non-English inputs \citep{shi2023multilingualcot,etxaniz2024english}. These conditions make low-resource emotion detection useful for testing whether multi-agent diversity creates genuinely complementary evidence or merely repeats shared cross-lingual failure modes.

\paragraph{Multi-agent emotion reasoning and our gap.}
The closest multi-agent emotion-reasoning work proposes specialized agents for perception, reasoning, and resolution. That line of work motivates modular affective reasoning, but it studies a fixed coordination design rather than separating topology from diversity source. Similarly, prompting work studies role variation without learned parameter diversity, while adapter work usually evaluates specialization under one aggregation regime. The gap is therefore not the absence of multi-call inference, prompts, or adapters; it is the absence of a controlled comparison that asks whether stochastic, prompted, and learned diversity should be aggregated in parallel or chained sequentially under the same inference budget.

\section{Methodology}
\label{sec:method}

\subsection{Task and Evaluation Setting}
\label{sec:task}

We formulate emotion detection as multi-label classification over the six BRIGHTER emotions: \textsc{anger}, \textsc{disgust}, \textsc{fear}, \textsc{joy}, \textsc{sadness}, and \textsc{surprise}. Let \(\mathcal{E}\) denote this label set, with \(|\mathcal{E}|=6\). Each example is a triple \((x_i,L_i,y_i)\), where \(x_i\) is the input text, \(L_i\) is the language identifier, and \(y_i\in\{0,1\}^{|\mathcal{E}|}\) is the gold multi-label emotion vector. Each system emits a JSON object containing one binary decision per emotion and a short reasoning field.

The headline metric is Macro-F1 averaged uniformly over languages and emotions. For language \(L\) and emotion \(e\), let \(\mathrm{F1}_{L,e}\) be the binary F1 score for emotion \(e\) on examples from language \(L\). We report
\[
\mathrm{MacroF1}
=
\frac{1}{|\mathcal{L}|\,|\mathcal{E}|}
\sum_{L\in\mathcal{L}}
\sum_{e\in\mathcal{E}}
\mathrm{F1}_{L,e}
\]
Uniform language averaging prevents languages with larger test sets from dominating the aggregate score and makes language-level failures visible.

We evaluate on the nine-language low-resource BRIGHTER subset: Naija Pidgin, Mandarin Chinese, Marathi, Moroccan Arabic, Tatar, Emakhuwa, Mozambican Portuguese, Zulu, and Indonesian. We use the standard splits: 11,214 public training instances pooled over the seven languages with public training data, 1,975 development instances, and 8,597 held-out test instances \citep{muhammad2025semeval11}. All systems are evaluated as a single pooled multilingual system. L1/L2 and the prompting baselines are inference-only, whereas L3 adapters are trained once per backbone on the pooled public training set; no model is selected or tuned separately per target language. Within each backbone, all matrix cells use the same output schema, reasoning-bank policy, and decision protocol. Additional dataset, split, and reasoning-bank details are given in Appendix~\ref{app:dataset_details}.

\subsection{Topology--Diversity Matrix}

We study two topologies: \emph{Width} (parallel aggregation) and \emph{Depth} (sequential refinement), with three diversity sources: stochastic sampling (L1), designed prompts (L2), and learned specialization (L3), yielding the six cells in Figure~\ref{fig:topology-diversity}.

We instantiate the full matrix independently on Qwen2.5-14B-Instruct and Llama-3.1-8B-Instruct, served with vLLM \citep{kwon2023vllm}. Within each backbone, every cell makes exactly three inference calls per input, uses top-\(p=0.95\), and shares the same output schema and decision protocol. Each call receives the same shared input package $u_i=(x_i,L_i,b_{L_i})$ where \(b_{L_i}\) is the language-specific reasoning bank for language \(L_i\). This fixed-budget design makes within-backbone comparisons interpretable: observed differences can be attributed to topology, diversity source or their interaction rather than to a different number of calls or model identity inside the matrix.

\paragraph{Width topology.} In Width, three agents receive the same input package \(u_i\) and predict independently. For diversity level \(\delta\in\{\mathrm{L1},\mathrm{L2},\mathrm{L3}\}\), the three agent predictions are \[\hat{y}^{(k)}_i=f^{\delta}_{k}(u_i),\qquad k\in\{1,2,3\},\] where \(\hat{y}^{(k)}_i\in\{0,1\}^{|\mathcal{E}|}\). For each emotion label, the final prediction is obtained by majority vote: \[\hat{y}^{W}_{i,e}=\mathbf{1}\!\left[\sum_{k=1}^{3}\hat{y}^{(k)}_{i,e}\ge 2\right],\qquad e\in\mathcal{E}.\] Thus, a label is emitted iff at least two of the three agents emit it. Width has no inter-agent conditioning or ordering, and the three outputs are symmetrically aggregated.

\paragraph{Depth topology.} In Depth, three layers are executed sequentially. Let \(z_i^{(\ell)}\) denote the full JSON output of layer \(\ell\), including both the binary emotion decisions and the reasoning field, and let \(\pi(z)\) extract the binary emotion vector from a JSON output. Layer 1 predicts from the input alone: $z_i^{(1)}=f^{\delta}_{1}(u_i;T_1)$. Layer 2 receives the original input together with Layer 1's full JSON output: $z_i^{(2)}=f^{\delta}_{2}(u_i,z_i^{(1)};T_2)$. Layer 3 receives the original input and the JSON outputs from both previous layers: $z_i^{(3)}=f^{\delta}_{3}(u_i,z_i^{(1)},z_i^{(2)};T_3)$. The final Depth prediction is $\hat{y}^{D}_i=\pi(z_i^{(3)})$. The temperature schedule descends as $(T_1,T_2,T_3)=(0.7,0.5,0.3)$, so that the amount of decoding randomness matches the layer's role in the chain. The first layer uses the highest temperature to allow alternative interpretations of ambiguous affective cues; the second layer reduces randomness while reviewing the first output; and the third layer uses the lowest temperature to make the final prediction more conservative and stable. This implements an exploration-to-consolidation pattern: early layers may explore alternative interpretations, while later layers are encouraged to revise and commit.

\subsection{Diversity Sources}

\paragraph{L1: Stochastic diversity.}
Calls share the generic CoT prompt and differ only through decoding stochasticity and, for Depth, sequential conditioning. Width-L1 uses seeds 42, 123, and 456 at \(T=0.7\); Depth-L1 uses the shared prompt under the \(0.7/0.5/0.3\) schedule. Seeds are fixed a priori rather than selected by development performance.

\paragraph{L2: Designed prompt diversity.}
Width-L2 uses three additive analytical lenses emphasizing surface, contextual, and discourse-structural cues at \(T=0.7\) and top-\(p=0.95\). Depth-L2 instead assigns sequential \textsc{Analyst}, \textsc{Critic}, and \textsc{Calibrator} roles. The prompts bias attention without restricting agents from using other evidence types; full templates are given in Appendices~\ref{app:width-prompts}--\ref{app:depth-prompts}.

\paragraph{L3: Learned specialization (parameter diversity).}
L3 trains three QLoRA adapters over each frozen backbone. Unlike L1/L2, L3 introduces pooled supervised adaptation in addition to inter-agent heterogeneity, so the raw L3--L1/L2 difference cannot be attributed to diversity alone. Section~\ref{sec:attribution} separates these effects using a shared-adapter control and a temperature-matched specialist control.

Width-L3 uses general, ambiguity, and contrastive specialists followed by the same majority vote used throughout Width. Depth-L3 uses position-specific analyst, critic, and calibrator adapters trained for successive correction. Adapters are hot-swapped through one vLLM engine; objectives and QLoRA details are provided in Appendix~\ref{app:adapter_training}.

\subsection{Shared Protocol and Ablations}
\label{sec:shared_protocol}

Within each backbone, all six cells share the same language-specific reasoning banks, precision-prior decision policy, JSON schema, and three-call budget. This prevents language conditioning, thresholding, or formatting differences from being mistaken for topology or diversity effects.

We test shared-adapter controls, temperature and prompt variants, specialist leave-one-out, alternative aggregation, paired-bootstrap uncertainty, cross-backbone replication, and Depth truncation. The complete ablation suite is reported in Appendix~\ref{app:ablations}.

\section{Experiments}
\label{sec:experiments}

We evaluate the complete six-cell matrix on the same nine-language BRIGHTER development and held-out test splits for Qwen2.5-14B-Instruct and Llama-3.1-8B-Instruct. Topology, prompts, reasoning banks, adapter objectives, output schema, and the three-call budget are unchanged across backbones.

The Qwen reference systems are zero-shot, few-shot, single-pass CoT, and seven-call CoT with self-consistency. Qwen2.5-72B and multilingual encoder results are reported only as contextual BRIGHTER references.

For key contrasts, we compute 95\% paired non-parametric bootstrap intervals using 2{,}000 shared resamples. For the learned Width--Depth comparison we additionally report a Wilcoxon signed-rank test over the nine paired language scores. Bootstrap intervals measure evaluation-sample uncertainty; targeted W-L3 seed perturbations separately assess sensitivity to training and inference stochasticity. Full compute and configuration details appear in Appendix~\ref{app:experimental_details}.

\section{Results}
\label{sec:results}
\subsection{Topology and Diversity Effects}

Table~\ref{tab:matrix} reports the complete Topology $\times$ Diversity matrix on both backbones. On Qwen2.5-14B, \wlthree reaches 53.61\% Macro-F1 on development and 52.83\% on test, and replacing designed prompt diversity (\wltwo, 40.71\%) with learned specialization improves test Macro-F1 by 12.12 points. The second backbone preserves the main diversity-source result. On Llama-3.1-8B, \wlthree reaches 52.94\% test Macro-F1, exceeding \wlone by 8.39 points and \wltwo by 7.85 points. Thus, the largest shifts in both matrices come from moving to the learned-specialization regime rather than from changing topology.

\begin{table*}[t]
\centering
\small
\begin{tabular}{llrrrr}
\toprule
\textbf{Topology} & \textbf{Diversity} & \multicolumn{2}{c}{\textbf{Qwen2.5-14B}} & \multicolumn{2}{c}{\textbf{Llama-3.1-8B}} \\
\cmidrule(lr){3-4}\cmidrule(lr){5-6}
 & & \textbf{Dev} & \textbf{Test} & \textbf{Dev} & \textbf{Test} \\
\midrule
\w & L1 Stochastic & $42.00$ & $40.66$ & $45.97$ & $44.55$ \\
\w & L2 Designed & $42.40$ & $40.71$ & $46.91$ & $45.09$ \\
\w & L3 Learned & $\mathbf{53.61}$ & $\mathbf{52.83}$ & $53.45$ & $\mathbf{52.94}$ \\
\dpth & L1 Stochastic & $44.61$ & $43.88$ & $48.84$ & $47.02$ \\
\dpth & L2 Designed & $45.69$ & $45.07$ & $48.52$ & $46.35$ \\
\dpth & L3 Learned & $50.74$ & $50.00$ & $\mathbf{53.78}$ & $52.77$ \\
\bottomrule
\end{tabular}
\caption{Topology $\times$ Diversity matrix averages over nine languages (Macro-F1, \%) on two model families/sizes. Values are the canonical runs; targeted seed robustness is reported in Appendix~\ref{app:seed_robustness}.}
\label{tab:matrix}
\end{table*}

The topology effect remains smaller and depends on the source of diversity. On Qwen, Depth outperforms Width for stochastic and designed diversity by 3.22 and 4.36 test points, while \wlthree exceeds \dlthree by 2.83 points. Llama shows the same directional topology advantage at L1 and L2, with Depth exceeding Width by 2.47 and 1.26 points, respectively, but the learned cells are nearly tied: \wlthree exceeds \dlthree by only 0.17 points. Therefore, no topology dominates across diversity sources, and the learned Width advantage observed on Qwen should not be treated as a backbone-independent rule.

The near-zero \wlone$\rightarrow$\wltwo improvement on Qwen is also informative, but it should not be read as evidence that designed prompts contain no useful signal. Under the canonical majority-vote rule, designed analytical-lens prompts add only 0.05 test points, from 40.66\% for \wlone to 40.71\% for \wltwo. However, the original aggregation ablation shows that \wltwo rises to 45.12\% under confidence-max aggregation, a gain of 4.41 points over majority vote (Appendix Table~\ref{tab:agg_appendix}). Prompt diversity therefore creates recoverable signal even when majority vote underuses it. The corresponding Llama Width result is likewise small under majority vote (45.09 versus 44.55), reinforcing that prompt-only diversity changes agent behavior much less than learned specialization.

This is especially plausible in the multilingual setting: surface, context, and structure lenses can redirect attention, but they cannot by themselves add missing language competence when the underlying model parameters remain unchanged. Learned specialization changes this regime by improving the competence and error profile of the individual agents before aggregation, as examined directly in Section~\ref{sec:attribution}.

\subsection{Cross-Backbone Replication}
\label{sec:cross_backbone}

The second backbone separates findings that replicate from those that are model-dependent. The learned-specialization advantage over stochastic and designed Width is large on both models: \wlthree$-$\wlone is +12.17 points on Qwen and +8.39 on Llama, while \wlthree$-$\wltwo is +12.12 and +7.85 points, respectively. Sequential conditioning also improves the stochastic cell on both backbones: \dlone$-$\wlone is +3.22 points on Qwen and +2.47 on Llama.

The learned Width--Depth ordering is less stable. On Qwen, the paired bootstrap difference \wlthree$-$\dlthree is +2.83 points with a 95\% interval of [1.91, 3.77]; on Llama it is +0.17 with an interval of [-0.57, 0.94]. A language-level Wilcoxon signed-rank test is also non-significant for this contrast on both Qwen ($p=.10$) and Llama ($p=.73$). We therefore retain the Qwen learned-Width result as a descriptive finding of that backbone, but the cross-backbone evidence supports the narrower conclusion that specialization effects replicate more strongly than a specific learned-topology preference.

\paragraph{Seed robustness.} We additionally perturb the stochastic components of the headline learned Width condition. On Qwen, changing the inference seed moves W-L3 test Macro-F1 from 52.83 to 53.38 ($\Delta=+0.55$); on Llama, an alternative QLoRA training seed moves it from 52.94 to 53.62 ($\Delta=+0.68$). These shifts are small relative to the 7.85--12.17 point learned-specialization gains over W-L1/W-L2. Appendix~\ref{app:seed_robustness} reports the corresponding robustness results.

\subsection{Comparison with Reference Systems}

Table~\ref{tab:baselines} compares the Qwen \wlthree result against the averaged reference systems used in our evaluation. The first group contains same-backbone no-multi-agent baselines. These are the primary Qwen comparisons because they share the Qwen2.5-14B-Instruct backbone and output protocol. The prompting baselines and L1/L2 cells are inference-only, whereas L3 uses pooled supervised adapter training; we therefore use these baselines to contextualize the gain rather than to attribute all of it to multi-agent diversity. \wlthree outperforms zero-shot prompting by 8.74 points, few-shot prompting by 8.60 points, single-pass CoT by 4.36 points and CoT+self-consistency at \(k=7\) by 4.02 points, while using three calls. This suggests that the gain is not simply from sampling more reasoning traces: CoT+self-consistency increases stochastic coverage around one prompt and one parameter setting, whereas \wlthree combines three learned specialists with different adapter priors before aggregation. The advantage is strongest as an averaged nine-language result rather than a claim of uniform dominance for every language and every comparator. In the matrix itself, \wlthree is the best cell on seven of the nine languages (Appendix Table~\ref{tab:perlang}); against SemEval contextual references, it exceeds the RoBERTa baseline on eight of nine languages but does not match the per-language Top-1 systems on most languages (Appendix Table~\ref{tab:semeval}).

\begin{table}[t]
\centering
\small
\begin{tabular}{lrrr}
\toprule
\textbf{Condition} & \textbf{Calls} & \textbf{Test F1} & \(\Delta\) \\
\midrule
Zero-shot Qwen2.5-14B & 1 & 44.09 & +8.74 \\
Few-shot Qwen2.5-14B & 1 & 44.23 & +8.60 \\
CoT, single pass & 1 & 48.47 & +4.36 \\
CoT+SC, \(k=7\) & 7 & 48.81 & +4.02 \\
\midrule
Qwen2.5-72B few-shot & 1 & 45.87 & +6.96 \\
XLM-R cross-lingual & -- & 33.27 & +19.56 \\
mDeBERTa cross-lingual & -- & 31.23 & +21.60 \\
\midrule
\wlthree (ours) & 3 & \textbf{52.83} & 0.00 \\
\bottomrule
\end{tabular}
\caption{Averaged reference systems on the nine-language test split (Macro-F1, \%). \(\Delta\) is the advantage of \wlthree over each condition. The first block contains same-backbone no-multi-agent baselines; the second block contains external BRIGHTER references.}
\label{tab:baselines}
\end{table}

The second group in Table~\ref{tab:baselines} contains BRIGHTER reference systems. \wlthree exceeds the Qwen2.5-72B few-shot reference by 6.96 points; this comparison is contextual because model scale, supervision, and inference protocol differ. It also exceeds the reported XLM-R and mDeBERTa cross-lingual encoder baselines by 19.56 and 21.60 points, respectively. These encoder comparisons are useful as context because they represent non-generative multilingual transfer systems from the same benchmark family. They should not be treated as controlled head-to-head evidence: the BRIGHTER references differ in supervision, model class, scale, and inference protocol, whereas the within-matrix and same-backbone Qwen comparisons remain the most interpretable evidence.

Appendix Table~\ref{tab:semeval} gives the per-language SemEval-2025 Task 11 comparison. We treat it as contextual rather than conclusive because leaderboard systems may use target-language supervision, per-language model selection and larger heterogeneous ensembles. \wlthree is higher than the official RoBERTa baseline on eight of nine languages, with an average advantage of 8.07 points; the main performance gap is in Marathi, where supervised RoBERTa is 10.00 points higher. We use \citet{dong2026emotion} as reference for multi-agent emotion reasoning rather than as a direct Macro-F1 baseline, since its English benchmarks, model scale and supervision setting differ from this study.

\subsection{Per-Language Patterns}

The gains from learned Width are not uniform across languages. Relative to \wltwo, the Qwen \wlthree improves Emakhuwa from 2.02\% to 19.16\% test Macro-F1 and Zulu from 13.69\% to 27.05\% (Appendix Table~\ref{tab:perlang}). These languages remain the lowest absolute scores, but the improvement is important because the prompt-only Width cells are dominated by all-zero abstention on the low-resource tail: prediction diagnostics show that \wlone and \wltwo emit all-zero vectors for roughly 97\% of Emakhuwa inputs and 86\% of Zulu inputs. Learned specialization therefore improves performance in a regime where prompt-only diversity mainly reproduces shared abstention.

The gains are not limited to these two lowest-resource languages. On Qwen, \wlthree exceeds \wlone on seven of nine languages; on Llama it exceeds \wlone on all nine (Appendix Tables~\ref{tab:perlang} and \ref{tab:perlang_llama}). Naija Pidgin, Moroccan Arabic, Tatar, Mozambican Portuguese, and Zulu show substantial gains on both backbones. Marathi remains an informative boundary case: the Qwen \wlthree result is 72.20\%, while the per-language RoBERTa contextual reference reaches 82.20\% (Appendix Table~\ref{tab:semeval}), illustrating the regime where target-language supervised encoder training can still dominate a unified pooled-adapter system.

A protocol-matched diagnostic analysis further tests whether the gain from learned specialization is related to base-model competence. Using the single-pass $T=0.7$ CoT score as a per-language competence proxy, the \wlthree$-$\wlone gain is negatively associated with competence on both backbones. The relation is strongest on Llama (Spearman $\rho=-0.767$, $p=.016$); Qwen shows the same direction ($\rho=-0.567$, $p=.112$). With only nine languages, this is supporting rather than definitive evidence, but it helps explain why BRIGHTER is diagnostic for the present question: the benchmark supplies a wide natural range of model competence while keeping the task and label space fixed.
\subsection{Attributing the W-L3 Gain}
\label{sec:attribution}

The raw L3 gain combines supervised adaptation, specialist heterogeneity, and decoding temperature. We therefore evaluate the staged control in Table~\ref{tab:adaptation_ladder}: \wlone, a shared adapted model with three stochastic calls at \(T=0.7\), three specialist adapters at the same temperature, and canonical \wlthree at \(T=0.3\).

\begin{table}[t]
\centering
\small
\begin{tabular}{lrr}
\toprule
\textbf{Condition} & \textbf{Qwen} & \textbf{Llama} \\
\midrule
\wlone & 40.66 & 44.55 \\
Shared adapter, 3 seeds ($T=.7$) & 50.04 & 49.08 \\
Three specialist adapters ($T=.7$) & 51.57 & 49.83 \\
Canonical \wlthree & \textbf{52.83} & \textbf{52.94} \\
\bottomrule
\end{tabular}

\caption{Staged test-set controls for the learned Width gain (Macro-F1, \%).}

\label{tab:adaptation_ladder}

\end{table}

Along this intervention path, Qwen gains \(+9.38\) points from shared adaptation, \(+1.53\) from specialist differentiation at matched temperature, and \(+1.26\) from the canonical lower temperature. The corresponding Llama increments are \(+4.53\), \(+0.75\), and \(+3.11\). Thus, the largest L3 improvement comes from supervised adaptation, with a smaller additional contribution from specialist heterogeneity. We therefore interpret L3 as a \emph{learned-specialization} regime rather than a pure diversity intervention.

Additional Qwen controls support the same interpretation: raising \wlthree to \(T=0.7\) reduces test Macro-F1 to 51.57, removing the system prompt costs about 2.8 points, and adding specialist-specific prompts produces no meaningful improvement (Appendix~\ref{app:ablations}).

The one-call adapted comparison shows that individual specialists are already strong. Qwen specialists score 52.42, 50.06, and 51.75 versus 52.83 for the ensemble; Llama specialists score 52.33, 47.50, and 53.52 versus 52.94. Aggregation therefore contributes modest, backbone-dependent value rather than explaining the full L3 gain. Qwen leave-one-out re-aggregation additionally drops by 5.29, 4.78, and 2.01 points when removing the general, contrastive, and ambiguity specialists, respectively (Appendix~\ref{app:ablations}).

\subsection{Learned Depth vs. Learned Width}

The L3 reversal on Qwen is the clearest topology--specialization interaction, but the second backbone shows that its aggregate magnitude is model-dependent. Table~\ref{tab:truncation} reports the Qwen truncation analysis. Depth itself is not generally harmful: for \dlone and \dltwo, the full chain improves over Layer 1 alone by 2.01 and 3.21 test points. The pattern reverses only for Qwen \dlthree, where Layer 1 alone already reaches 52.11\% and the full learned chain falls to 50.00\%. This supports an implementation-specific over-correction interpretation rather than a general weakness of sequential refinement.

The aggregate degradation does not repeat on Llama: the learned Depth Layer-1 output is 52.33\%, while the full chain reaches 52.77\%. Nevertheless, the decision-transition analysis exposes a related over-revision tendency. From Layer 1 to Layer 2, Qwen fixes 2.47\% of label decisions but breaks 3.46\% (net $-0.99$ points at the decision level); Llama fixes 1.05\% but breaks 2.46\% (net $-1.42$). Thus, correction-trained later specialists overwrite correct early decisions on both backbones, but whether those changes reduce the final language-averaged Macro-F1 depends on the model and language distribution.

The Qwen skip-layer variant is also informative: removing Layer 2 scores 50.04\%, nearly identical to the full \dlthree chain at 50.00\%. Layer 2 is therefore not the unique bottleneck in Qwen. The combined evidence supports a narrow practical conclusion: sequential learned specialists need preservation-aware, position-aware calibration so that later layers learn when \emph{not} to revise a strong predecessor. The data do not support a universal rule that learned specialists should always be run in parallel.

Post-hoc aggregation remains a diagnostic factor for prompt-only Width. In the Qwen aggregation run, replacing majority vote with confidence-max improves \wlone from 40.66\% to 44.39\% and \wltwo from 40.71\% to 45.12\%, gains of 3.73 and 4.41 points, respectively. In contrast, \wlthree is comparatively stable under that readout. Together with the paired cross-backbone statistics in Section~\ref{sec:cross_backbone}, these results reinforce the main distinction: learned specialization is the robust large effect, while topology and aggregation effects are smaller and conditional on how agent differences are created.

\section{Conclusion}

We set out to ask whether topology or diversity source matters more in multi-agent LLM inference. The central pattern is consistent: under a fixed three-call budget, changing how agents are differentiated produces substantially larger performance shifts than changing topology. On Qwen, moving from designed prompts to learned QLoRA specialists gains 12.12 Macro-F1 points; the corresponding gain on Llama is 7.85 points. The cross-backbone evidence, however, qualifies the topology claim. Sequential depth helps the stochastic and prompted regimes, while the learned Width--Depth ordering is clear on Qwen but nearly tied on Llama. The attribution controls further show that the L3 advantage is driven primarily by pooled supervised adaptation, with a smaller additional contribution from specialist heterogeneity. The practical upshot is therefore to invest first in how agents acquire complementary competence, and to evaluate topology jointly with that specialization rather than treating Width or Depth as universally preferable. Reporting multi-agent systems along these axes separately would make gains in this fast-moving area considerably easier to interpret and build on.

\section*{Limitations}
The matrix is evaluated on two decoder-only instruction-tuned backbones, Qwen2.5-14B-Instruct and Llama-3.1-8B-Instruct, which does not establish model-independent design rules across architectures or scales. The study uses one task domain, BRIGHTER multilingual multi-label emotion detection; its variation in language competence is useful diagnostically, but the observed topology--specialization interactions may not transfer unchanged to unrelated agent tasks. L3 also differs from L1/L2 in supervision: the learned adapters use pooled public training data from the seven Track-A languages. The target-excluded reasoning-bank policy depends on imperfect proxies: Chinese and Tatar lack close in-pool siblings and, therefore, rely on broader cues such as English pretraining coverage, script overlap, contact, and emoji anchors. Paired bootstrap intervals quantify evaluation-sample uncertainty rather than training-run variance; targeted W-L3 seed perturbations provide a separate sensitivity check and shift test Macro-F1 by 0.55 points on Qwen and 0.68 points on Llama. Emakhuwa also remains low even in the best Qwen cell (19.16\% Macro-F1), showing that multi-agent inference cannot replace missing target-language competence.

\section*{Ethical considerations}
Emotion detection can support accessibility and crisis-monitoring applications, but it can also enable surveillance, profiling or unfair moderation. These risks are amplified in low-resource languages, where errors are higher and cultural cues are easier to misread. The system should not be used as an individual decision-maker in high-stakes settings; deployment requires human review, uncertainty reporting and validation with language-community expertise. We use publicly released BRIGHTER annotations, collect no new human-subject data, report aggregate results, and avoid reproducing identifiable examples. AI-assisted tools were used for language and LaTeX/submission support; the authors checked all experiments, numerical results, and scientific claims. We use BRIGHTER in accordance with the dataset's stated license and intended research use.

\section*{Acknowledgments}
We thank the anonymous ARR reviewers for constructive feedback that strengthened the cross-backbone and robustness analyses.

\bibliography{references}

\appendix

\section{Extended Related Work}
\label{app:extended_related}

This appendix expands the shortened Related Work in Section~\ref{sec:related}. The main paper keeps only the claims needed to motivate the topology--diversity matrix; the additional background here documents the broader literature connections. References for the methods summarized below are given in Section~\ref{sec:related}; the additional prompt-sensitivity reference is cited locally.

\paragraph{Parallel and sequential multi-agent inference.}
Parallel multi-call methods usually generate several candidate answers and aggregate them. Self-consistency samples multiple Chain-of-Thought traces from the same prompt and marginalizes over the final answers. Debate-style systems introduce interaction among agents, allowing one model or role to critique another before a final answer is chosen. Mixture-of-agents systems aggregate outputs from multiple proposers, sometimes across heterogeneous models or layers. Sequential methods instead refine a single trajectory: Self-Refine separates generation from feedback and revision, while Reflexion adds evaluative reflection across attempts. These systems demonstrate the usefulness of multiple calls, but they typically change topology, role structure, and diversity source together.

\paragraph{Why topology and diversity must be separated.}
The distinction matters because an additional call can help for different reasons. In a parallel topology, independent errors may cancel under aggregation. In a sequential topology, later calls may add evidence or correct earlier mistakes. However, the same topology can behave differently depending on how the calls are diversified. If agents differ only by sampling seed, their errors may remain correlated. If they differ by role prompt, they may attend to different evidence but still share the same underlying model limitations. If they differ by trained adapter weights, they may become individually stronger but may also become overconfident specialists. Our matrix separates these possibilities by crossing topology with stochastic, prompted, and learned diversity under the same three-call budget, instantiated independently on two backbones.

\paragraph{Diversity and ensemble competence.}
Ensemble theory suggests that aggregation is useful only when individual members are both competent and meaningfully diverse. Diversity alone is not sufficient: if all members share the same systematic error, a majority vote can simply reproduce that error. This is especially relevant for multilingual low-resource inference, where all calls may inherit the same weak language competence from the shared backbone. The learned-diversity condition is therefore not only a diversity intervention but also a competence intervention: QLoRA specialists can change the effective parameters used by each call while still preserving a shared base model.

\paragraph{Prompted diversity and role design.}
Role prompts and persona prompts can provide useful cognitive scaffolding, but they are sensitive to wording and may degrade performance when they over-constrain the model \citep{sclar2024prompt}. This motivates our additive lens design. The Surface, Context, and Structure agents are asked to pay particular attention to different evidence types, but they are not instructed to ignore the others. The goal is to bias attention without reducing individual competence, since restrictive role prompts can lower the quality of the members that aggregation depends on.

\paragraph{Multilingual transfer and low-resource emotion detection.}
Multilingual emotion detection exposes topology--diversity interactions that English-only evaluation may hide. Cross-lingual transfer is affected by typological similarity, script overlap, tokenization, pretraining coverage, and resource level. Recent work further suggests that multilingual LLM reasoning may route through English and transfer unevenly across target languages. In this setting, stochastic and prompt-induced diversity may vary reasoning style without creating genuinely independent evidence for the lowest-resource languages. This motivates the learned-adapter condition and the per-language reporting used throughout the paper.

\section{Dataset, Split, and Reasoning-Bank Details}
\label{app:dataset_details}
\label{app:reasoning-banks}

This section expands the evaluation-setting details compressed from Section~\ref{sec:task}. We evaluate on the nine-language low-resource BRIGHTER subset: Naija Pidgin (\texttt{pcm}), Mandarin Chinese (\texttt{chn}), Marathi (\texttt{mar}), Moroccan Arabic (\texttt{ary}), Tatar (\texttt{tat}), Emakhuwa (\texttt{vmw}), Mozambican Portuguese (\texttt{ptmz}), Zulu (\texttt{zul}), and Indonesian (\texttt{ind}) \citep{muhammad2025brighter}. These languages span five scripts: Latin, Arabic, Chinese characters, Devanagari, and Cyrillic. They also cover several language families, including Niger-Congo, Sino-Tibetan, Indo-European, Turkic, Afro-Asiatic, and Austronesian.

We use the standard BRIGHTER splits: 11{,}214 public training instances pooled over the seven languages with public training data, 1{,}975 development instances, and 8{,}597 held-out test instances. Under the BRIGHTER/SemEval protocol, Zulu and Indonesian are Track-C languages with no target-language training rows. We therefore train learned adapters only on the pooled public training data from the seven Track-A languages and evaluate the resulting single multilingual system on all nine languages.

The reasoning-bank policy is target-excluded. Same-family sources are used when available; otherwise the bank falls back to script overlap, sociolinguistic contact, pretraining coverage, or script-agnostic affect anchors such as emoji. The purpose of the reasoning bank is not to provide target-language labeled supervision, but to supply compact language-specific context, cue reminders, and affect anchors that are held fixed across all topology--diversity cells.

\subsection{Design evolution}
\label{app:bank-evolution}

The current bank design is the fourth iteration in an internal sequence. V1 attached one-line rationales directly to few-shot exemplars. This was auditable but inflated prompt length and made every exemplar edit require a paired rationale edit. V2 moved the rationale into a single shared reasoning seed per language, but this was too under-specified for languages with diverse affective cues. V3 expanded to twelve seeds per language, which improved coverage but increased context length and encouraged verbose reasoning. V4, the canonical design, uses four cue-first reasoning seeds per language. This was selected as a coverage-versus-brevity compromise and is reused unchanged across all six matrix cells.

\subsection{Bank JSON schema}
\label{app:bank-schema}

Table~\ref{tab:bank-schema} shows the schema for each language reasoning bank. The same file format is used for all nine languages.

\begin{table}[t]
\centering\small
\begin{tabular}{p{0.92\columnwidth}}
\toprule
Reasoning bank file: \texttt{\{lang\}\_reasoning\_seed42.json} \\
\midrule
\texttt{\{} \\
\texttt{~~"language\_name": "\{full name\}",} \\
\texttt{~~"script": "\{Latin | Arabic | Cyrillic | Devanagari | Han\}",} \\
\texttt{~~"family": "\{family or proxy class\}",} \\
\texttt{~~"seed": 42,} \\
\texttt{~~"language\_instructions": "\{one-paragraph cue-and-context primer\}",} \\
\texttt{~~"reasoning\_examples": [} \\
\texttt{~~~~\{} \\
\texttt{~~~~~~"id": "\{example id\}", "emotion": "\{emotion label\}",} \\
\texttt{~~~~~~"text": "\{exemplar text\}",} \\
\texttt{~~~~~~"reasoning": \{} \\
\texttt{~~~~~~~~"cues": "\{lexical and emoji evidence\}",} \\
\texttt{~~~~~~~~"tone": "\{discourse-level affect\}",} \\
\texttt{~~~~~~~~"interpretations": "\{alternative readings\}",} \\
\texttt{~~~~~~~~"critique": "\{adjudication leading to label\}"} \\
\texttt{~~~~~~\},} \\
\texttt{~~~~~~"final\_labels": ["\{emotion label\}"]} \\
\texttt{~~~~\}, ...} \\
\texttt{~~]} \\
\texttt{\}} \\
\bottomrule
\end{tabular}
\caption{Reasoning bank file schema. One file is used per target language under \texttt{data/reasoning\_banks/}. The rendered bank is inserted into the CoT block in Table~\ref{tab:cot-block}.}
\label{tab:bank-schema}
\end{table}

\subsection{Family/proxy transfer policy}
\label{app:proxy-policy}

The reasoning banks follow a target-excluded family/proxy policy. Same-family sources are used when available; otherwise, the bank falls back on script overlap, sociolinguistic contact, pretraining coverage, or script-agnostic affect anchors such as emoji. Table~\ref{tab:proxy-policy} summarizes the policy used for each target language. This table is included to make clear that proxy cases such as Chinese and Tatar are treated as bounded approximations, not as typological matches.

\begin{table}[t]
\centering\small
\setlength{\tabcolsep}{4pt}
\begin{tabular}{l l l p{0.30\columnwidth}}
\toprule
\textbf{Target} & \textbf{Family} & \textbf{Script} & \textbf{Proxy basis} \\
\midrule
\texttt{pcm}  & Creole & Latin & Native Pidgin examples plus English-lexifier and substrate cues \\
\texttt{chn}  & Sino-Tibetan & Han & English-dominant proxy with script-agnostic emoji anchors \\
\texttt{mar}  & Indo-European & Devanagari & Broad Indo-European pool, including Hindi as the closest sibling \\
\texttt{ary}  & Afro-Asiatic & Arabic & Maghrebi Arabic sibling and Afro-Asiatic proxies \\
\texttt{tat}  & Turkic & Cyrillic & Russian/Ukrainian proxies via script overlap and contact \\
\texttt{vmw}  & Niger-Congo & Latin & Niger-Congo and Bantu-family proxies \\
\texttt{ptmz} & Indo-European & Latin & Romance and wider Indo-European proxies \\
\texttt{zul}  & Niger-Congo & Latin & Niger-Congo and Bantu-family proxies \\
\texttt{ind}  & Austronesian & Latin & Austronesian sibling and regional-register cues \\
\bottomrule
\end{tabular}
\caption{Family/proxy basis for each of the nine target languages. The policy prioritizes same-family target-excluded sources when available and uses proxy-transfer fallbacks where the available pool lacks close siblings.}
\label{tab:proxy-policy}
\end{table}

\section{Experimental Details}
\label{app:experimental_details}

This section expands the compressed experimental protocol in Section~\ref{sec:experiments}. The complete six-cell matrix is evaluated on the same BRIGHTER development and held-out test splits for Qwen2.5-14B-Instruct and Llama-3.1-8B-Instruct. Within each backbone, cells share the same language-specific reasoning banks, JSON output schema, precision-prior decision policy, and three-call inference budget. Llama is used as the second-backbone replication of the matrix.

\subsection{Comparison groups}
\label{sec:comparison_groups}
The baselines use the same Qwen2.5-14B-Instruct backbone and the same output protocol as the Qwen matrix cells: zero-shot prompting, few-shot prompting, single-pass CoT, and seven-call CoT with self-consistency. These are the primary comparisons because differences are attributable to inference design rather than backbone, label schema, or output format. We additionally report BRIGHTER reference systems, including Qwen2.5-72B few-shot prompting, XLM-R, and mDeBERTa, as contextual scale and cross-lingual references. SemEval-2025 leaderboard comparisons are reported separately in Appendix~\ref{app:semeval}, but are not treated as important because those systems may use target-language supervision, per-language tuning, or larger ensembles.

\subsection{Statistical reporting}
Table~\ref{tab:matrix} reports the canonical run for each matrix cell. Bootstrap intervals quantify uncertainty over evaluation examples, while Appendix~\ref{app:seed_robustness} reports targeted W-L3 training/inference seed perturbations as a separate robustness check. For key contrasts, we compute 95\% paired non-parametric percentile bootstrap confidence intervals using 2{,}000 replicates. Each replicate resamples examples within language using the same draw for both systems and recomputes the language-uniform Macro-F1 difference. For the learned Width--Depth contrast we also report the Wilcoxon signed-rank test over the nine paired language-level scores as a complementary cross-language consistency check.

\subsection{Compute reporting}
Table~\ref{tab:compute_appendix} reports the per-cell development-split compute budget. All matrix cells issue the same number of requests, while token counts and wall-clock time vary slightly with topology and prompt length. These numbers report inference only and exclude one-time QLoRA adapter training. Across the Qwen matrix, inference used approximately 73.6 GPU-hours on development and 155 GPU-hours on test, excluding one-time adapter training.

\begin{table}[t]
\centering
\small
\begin{tabular}{lrrr}
\toprule
\textbf{Cell} & \textbf{Requests} & \textbf{Tokens (M)} & \textbf{Wall (h)} \\
\midrule
\wlone & 5{,}925 & 6.74 & 11.7 \\
\wltwo & 5{,}925 & 7.26 & 12.8 \\
\wlthree & 5{,}925 & 6.46 & 11.8 \\
\dlone & 5{,}925 & 7.84 & 12.3 \\
\dltwo & 5{,}925 & 8.21 & 12.9 \\
\dlthree & 5{,}925 & 7.66 & 12.1 \\
\bottomrule
\end{tabular}
\caption{Per-cell Qwen2.5-14B inference compute on the development split using a single 40GB GPU.}
\label{tab:compute_appendix}
\end{table}

\section{Per-Language Matrix Detail}
\label{app:perlang}

Tables~\ref{tab:perlang} and \ref{tab:perlang_llama} expand the averaged matrix results from Table~\ref{tab:matrix} into per-language test Macro-F1 for Qwen and Llama. The tables show that the learned-specialization gain is not driven by a single language and also expose the substantial remaining variation across languages.

\begin{table*}[t]
\centering
\small
\begin{tabular}{lrrrrrrrrrr}
\toprule
\textbf{Cell} & \textbf{pcm} & \textbf{chn} & \textbf{mar} & \textbf{ary} & \textbf{tat} & \textbf{vmw} & \textbf{ptmz} & \textbf{zul} & \textbf{ind} & \textbf{Avg} \\
\midrule
\wlone & 35.65 & 59.53 & 74.56 & 44.71 & 43.43 & 4.39 & 38.60 & 13.59 & 51.48 & 40.66 \\
\wltwo & 36.69 & 59.96 & 73.70 & 46.59 & 41.46 & 2.02 & 40.27 & 13.69 & 52.00 & 40.71 \\
\wlthree & 57.02 & 67.12 & 72.20 & 62.44 & 63.02 & 19.16 & 56.40 & 27.05 & 51.03 & 52.83 \\
\dlone & 40.43 & 61.36 & 76.30 & 51.21 & 47.52 & 6.39 & 41.79 & 15.80 & 54.11 & 43.88 \\
\dltwo & 43.18 & 60.57 & 76.68 & 50.49 & 48.23 & 7.74 & 45.13 & 18.96 & 54.66 & 45.07 \\
\dlthree & 47.52 & 64.40 & 75.76 & 56.92 & 55.23 & 18.55 & 52.15 & 26.78 & 52.67 & 50.00 \\
\bottomrule
\end{tabular}
\caption{Per-language Qwen2.5-14B test Macro-F1 (\%) for every matrix cell; Avg is the uniform mean over the nine languages.}
\label{tab:perlang}
\end{table*}

\begin{table*}[t]
\centering
\small
\begin{tabular}{lrrrrrrrrrr}
\toprule
\textbf{Cell} & \textbf{pcm} & \textbf{chn} & \textbf{mar} & \textbf{ary} & \textbf{tat} & \textbf{vmw} & \textbf{ptmz} & \textbf{zul} & \textbf{ind} & \textbf{Avg} \\
\midrule
\wlone & 38.51 & 60.14 & 78.42 & 45.95 & 49.23 & 10.28 & 51.03 & 16.53 & 50.88 & 44.55 \\
\wltwo & 39.64 & 60.29 & 77.64 & 46.79 & 50.63 & 10.95 & 51.44 & 16.58 & 51.86 & 45.09 \\
\wlthree & 55.87 & 64.77 & 79.45 & 55.64 & 60.65 & 19.47 & 57.55 & 30.45 & 52.63 & 52.94 \\
\dlone & 43.50 & 61.47 & 77.94 & 49.98 & 48.20 & 14.73 & 49.15 & 22.34 & 55.86 & 47.02 \\
\dltwo & 44.23 & 60.46 & 78.58 & 48.46 & 46.31 & 13.86 & 48.93 & 20.59 & 55.72 & 46.35 \\
\dlthree & 54.53 & 65.39 & 80.63 & 54.10 & 58.92 & 17.68 & 58.49 & 30.46 & 54.71 & 52.77 \\
\bottomrule
\end{tabular}
\caption{Per-language Llama-3.1-8B-Instruct test Macro-F1 (\%) for every matrix cell; Avg is the uniform mean over the nine languages.}
\label{tab:perlang_llama}
\end{table*}

\section{Seed Robustness}
\label{app:seed_robustness}

This appendix reports targeted seed perturbations for the headline W-L3 condition. The canonical run remains the value reported in Table~\ref{tab:matrix}; these perturbations are a sensitivity check rather than a full-matrix estimate of training-run variance. On Qwen2.5-14B, an alternative inference seed produces 54.70 dev and 53.38 test Macro-F1, versus 53.61 and 52.83 canonically. On Llama-3.1-8B, an alternative QLoRA training seed produces 54.73 dev and 53.62 test, versus 53.45 and 52.94 canonically. A further Llama seed-diversity variant reaches 53.67 dev and 52.89 test.

\begin{table}[t]
\centering
\small
\resizebox{\columnwidth}{!}{%
\begin{tabular}{llrrr}
\toprule
\textbf{Backbone} & \textbf{Perturbation} & \textbf{Canonical} & \textbf{Repeat} & $|\Delta|$ \\
\midrule
Qwen2.5-14B & Inference seed & 52.83 & 53.38 & 0.55 \\
Llama-3.1-8B & Training seed & 52.94 & 53.62 & 0.68 \\
Llama-3.1-8B & Seed-diversity variant & 52.94 & 52.89 & 0.05 \\
\bottomrule
\end{tabular}%
}
\caption{Individual W-L3 seed runs on the test split (Macro-F1, \%). The observed shifts are much smaller than the learned-specialization gains over W-L1 and W-L2.}
\label{tab:seed_robustness}
\end{table}

\section{SemEval Reference Detail}
\label{app:semeval}

Table~\ref{tab:semeval} provides the per-language reference comparison used in Section~\ref{sec:results}. These numbers are contextual rather than primary evidence because SemEval systems may use target-language supervision, per-language model selection, and larger ensembles. The table shows that \wlthree improves over the RoBERTa reference on eight of nine languages, while still trailing the per-language Top-1 systems on most languages.

\begin{table*}[t]
\centering
\small
\begin{tabular}{llrrrrr}
\toprule
\textbf{Lang} & \textbf{Track} & \textbf{\wlthree} & \textbf{RoBERTa} & $\Delta_R$ & \textbf{Top-1} & $\Delta_1$ \\
\midrule
pcm & A & 0.570 & 0.555 & +0.015 & 0.674 & -0.104 \\
chn & A & 0.671 & 0.531 & +0.140 & 0.709 & -0.038 \\
mar & A & 0.722 & 0.822 & -0.100 & 0.884 & -0.162 \\
ary & A & 0.624 & 0.472 & +0.152 & 0.629 & -0.005 \\
tat & A & 0.630 & 0.539 & +0.091 & 0.846 & -0.216 \\
vmw & A & 0.192 & 0.121 & +0.071 & 0.325 & -0.133 \\
ptmz & A & 0.564 & 0.459 & +0.105 & 0.548 & +0.016 \\
zul & C & 0.271 & 0.153 & +0.118 & 0.397 & -0.126 \\
ind & C & 0.510 & 0.376 & +0.134 & 0.672 & -0.162 \\
\midrule
Avg & -- & 0.528 & 0.448 & +0.080 & 0.632 & -0.104 \\
\bottomrule
\end{tabular}
\caption{Per-language comparison with SemEval-2025 references. $\Delta_R$ is \wlthree minus the RoBERTa baseline; $\Delta_1$ is \wlthree minus the per-language Top-1 submission.}
\label{tab:semeval}
\end{table*}

\section{Ablation Details}
\label{app:ablations}

This section collects the ablations supporting the mechanism claims in Section~\ref{sec:results}. The ablations test whether the \wlthree gain is due to learned adapters rather than co-varying temperature or prompt choices, whether the \dlthree deficit is caused by depth itself or by learned-specialist interaction, and whether aggregation or bootstrap uncertainty changes the interpretation.

\begin{table*}[t]
\centering
\small
\resizebox{\textwidth}{!}{
\begin{tabular}{p{2.7cm}p{2.5cm}p{3.6cm}p{4.2cm}p{4.0cm}}
\toprule
Claim tested & Ablation(s) & Intervention & Main result & Interpretation \\
\midrule
W-L3 gain is mainly due to learned adapters, not decoding temperature &
A1 &
Run W-L3 at \(T=0.7\), matching the higher prompt-only sampling temperature &
W-L3 remains strong; increasing temperature reduces test Macro-F1 by 1.26 points relative to canonical W-L3 &
The learned-adapter gain is not explained by the lower \(T=0.3\) decoding setting. \\

W-L3 gain is not primarily caused by prompt engineering &
A7, A8 &
Replace the full system prompt with a minimal prompt, or remove the system prompt entirely &
Minimal prompt costs 2.71 points; no-system prompt costs 2.84 points, while W-L3 remains far above W-L2 &
Adapters carry most of the performance gain; prompt content contributes but is not the main driver. \\

Explicit per-agent prompts do not add to learned adapter diversity &
A9 &
Add per-agent prompts on top of the three W-L3 adapters &
Performance is essentially flat: \(+0.08\) dev and \(-0.12\) test &
Prompt diversity and adapter diversity do not compose additively in this setting. \\

All W-L3 specialists are important &
A2 &
Leave-one-out re-aggregation over the three Width-L3 adapters &
Removing \texttt{general}, \texttt{contrastive}, and \texttt{ambiguity} drops test Macro-F1 by 5.29, 4.78, and 2.01 points, respectively &
The Width-L3 ensemble is not driven by a single adapter; learned diversity is functionally used by aggregation. \\

Learned Depth underperforms because of sequential specialist interaction, not because depth is generally harmful &
A3, A13 &
Re-score Depth predictions after truncating at Layer 1 or Layer 2 across D-L1, D-L2, and D-L3 &
Full chains help D-L1 and D-L2, but D-L3 peaks at Layer 1: 52.11 versus 50.00 for the full chain &
The learned-depth failure is specific to trained specialists revising one another, not to JSON-conditioned depth itself. \\

Layer 2 is not the sole cause of D-L3 degradation &
A6 &
Skip Layer 2 in D-L3 and run Layer 1 directly into Layer 3 &
Skip-L2 scores 50.04, nearly identical to canonical D-L3 at 50.00 &
The degradation is caused by the later learned calibration process as a whole, not only by the critic layer. \\

Width aggregation rule is a diagnostic factor, especially for prompt-only systems &
A5 &
Re-aggregate saved Width predictions with confidence-max and any-of-three rules &
Confidence-max improves W-L1 by 3.73 and W-L2 by 4.41 points; W-L3 is comparatively stable, with any-of-three adding 1.13 points &
Prompt-only Width contains signal that majority vote underuses; W-L3 is less dependent on aggregation choice. \\

Headline learned-cell comparisons should be read with uncertainty estimates &
A4 &
Bootstrap test predictions with 2{,}000 paired resamples &
The W-L3 vs. W-L2 gap is much larger than the bootstrap uncertainty; the W-L3 vs. D-L3 gap is smaller and should be read more cautiously &
The main learned-diversity claim is robust, while the topology gap at L3 is directionally meaningful but less decisive. \\
\bottomrule
\end{tabular}
}
\caption{Ablation summary. Each ablation targets a specific mechanism required by the topology--diversity interpretation: whether W-L3 gains come from adapter weights rather than co-varying prompt or temperature choices, whether the D-L3 deficit is caused by depth itself or by learned specialist interaction, and whether aggregation or bootstrap uncertainty changes the interpretation.}
\label{tab:ablation_summary}
\end{table*}

\begin{table}[t]
\centering
\small
\begin{tabular}{lrrrr}
\toprule
\textbf{Cell} & \textbf{L1 only} & \textbf{L1+L2} & \textbf{Full} & \(\Delta\) \textbf{L1--Full} \\
\midrule
\dlone & 41.87 & 43.69 & \textbf{43.88} & -2.01 \\
\dltwo & 41.86 & \textbf{45.46} & 45.07 & -3.21 \\
\dlthree & \textbf{52.11} & 49.94 & 50.00 & +2.11 \\
\bottomrule
\end{tabular}
\caption{Depth truncation results on the test split (Macro-F1, \%). For D-L1 and D-L2, sequential refinement improves over Layer~1 alone. For D-L3, Layer~1 alone is strongest, showing that the learned-depth failure is specific to specialist interaction rather than depth itself.}
\label{tab:truncation}
\end{table}

\section{Additional Figures}
\label{app:figures}

\begin{figure*}[t]
  \centering
  \includegraphics[width=0.98\textwidth]{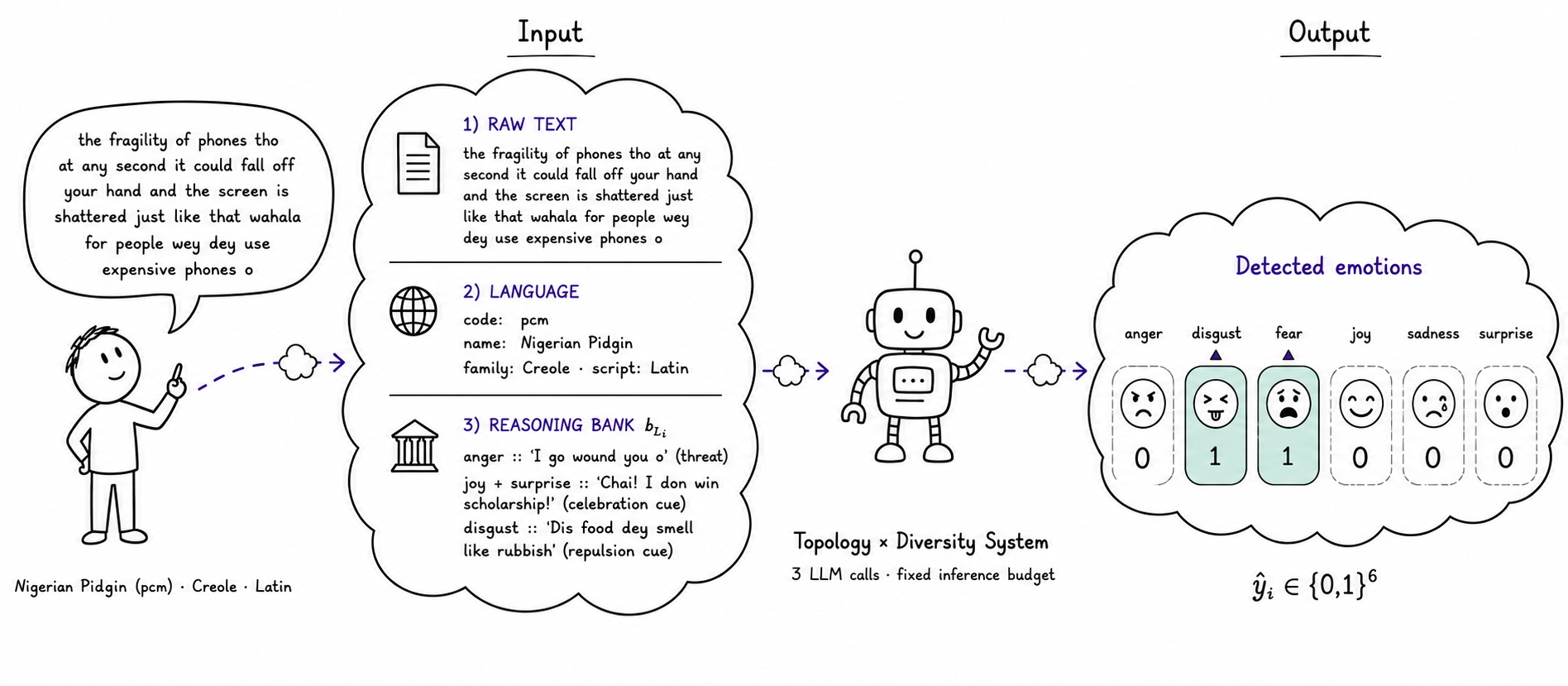}
  \caption{Input--output example for the shared multilingual emotion-detection interface. The input is a Nigerian Pidgin (\texttt{pcm}) post. Each call receives the raw text \(x_i\), language metadata \(L_i\), and the rendered reasoning bank \(b_{L_i}\); the topology--diversity system then connects and diversifies three LLM calls under the fixed inference budget. The output is a JSON-compatible multi-label prediction vector \(\hat{y}_i\in\{0,1\}^{6}\), ordered as \textsc{anger}, \textsc{disgust}, \textsc{fear}, \textsc{joy}, \textsc{sadness}, and \textsc{surprise}. The shown prediction activates \textsc{disgust} and \textsc{fear}.}
  \label{fig:input_output_example}
\end{figure*}

\begin{figure*}[t]
  \centering
  \includegraphics[width=0.92\textwidth]{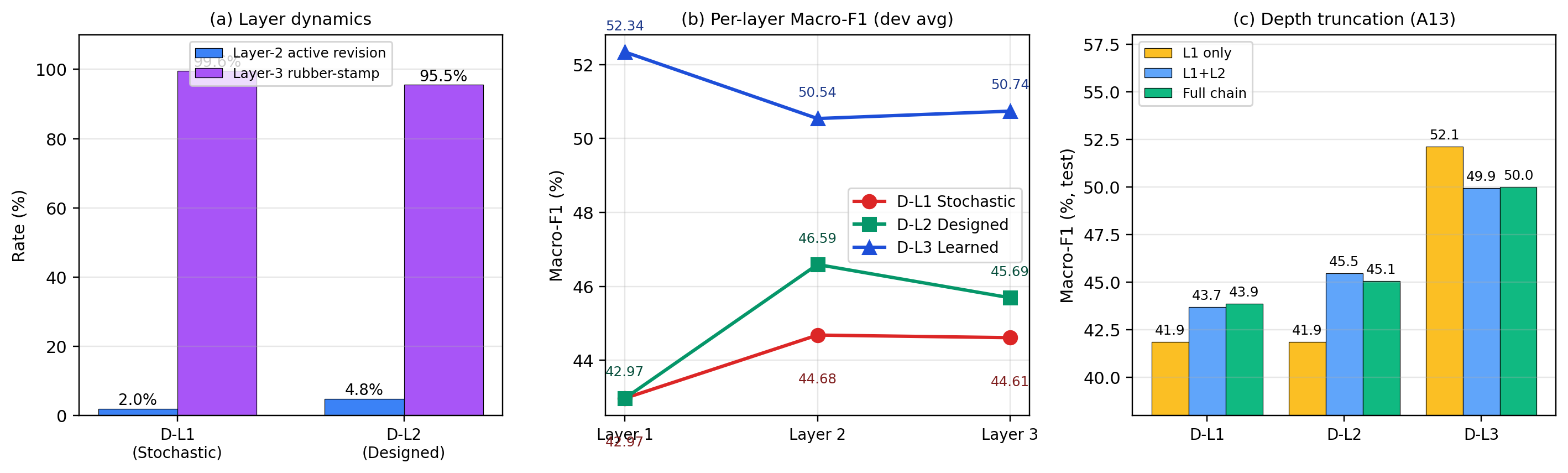}
  \caption{Depth dynamics and truncation analysis. D-L1 and D-L2 gain from sequential layers; D-L3 peaks at Layer 1 and erodes under later learned specialists.}
  \label{fig:depth_appendix}
\end{figure*}

\begin{figure*}[t]
  \centering
  \includegraphics[width=0.92\textwidth]{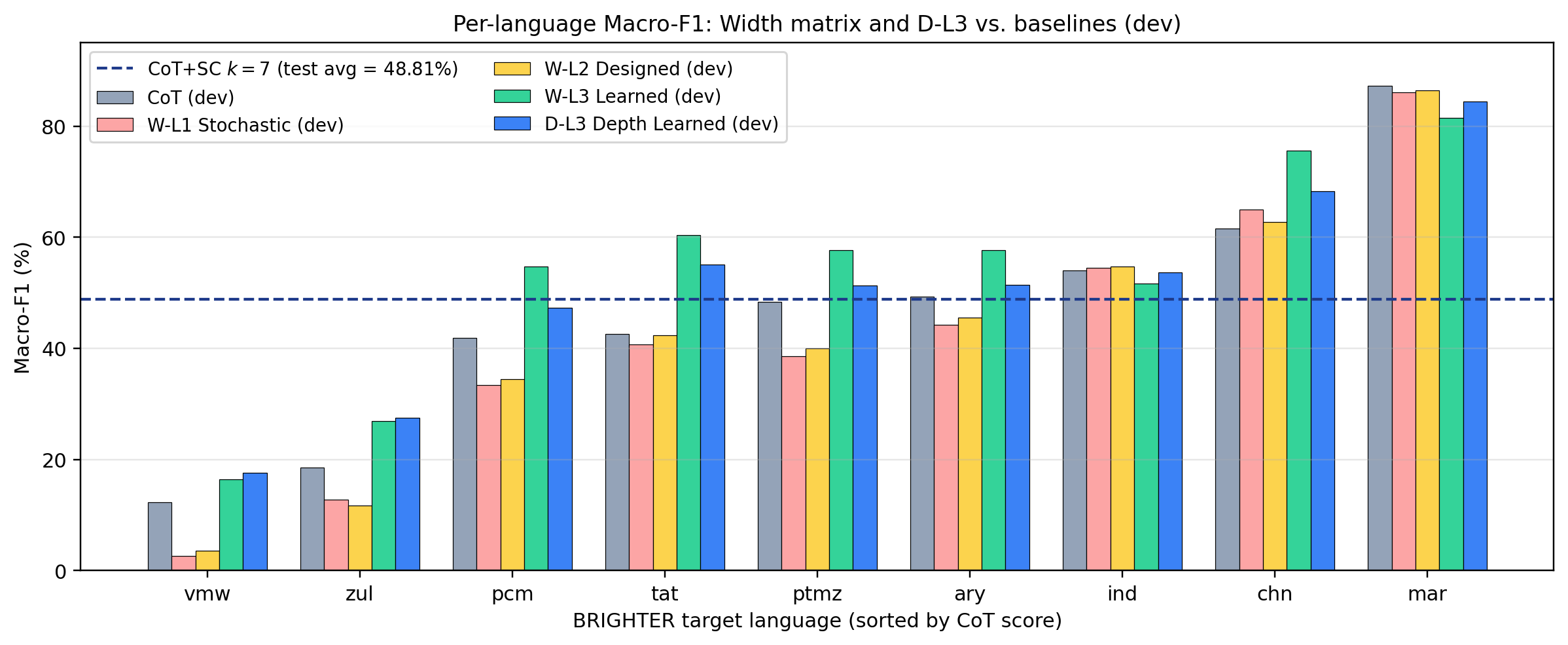}
  \caption{Per-language development results ordered by single-pass CoT score. The largest \wlthree gains appear on the low-resource tail.}
  \label{fig:perlang_appendix}
\end{figure*}

\begin{figure*}[t]
  \centering
  \includegraphics[width=0.92\textwidth]{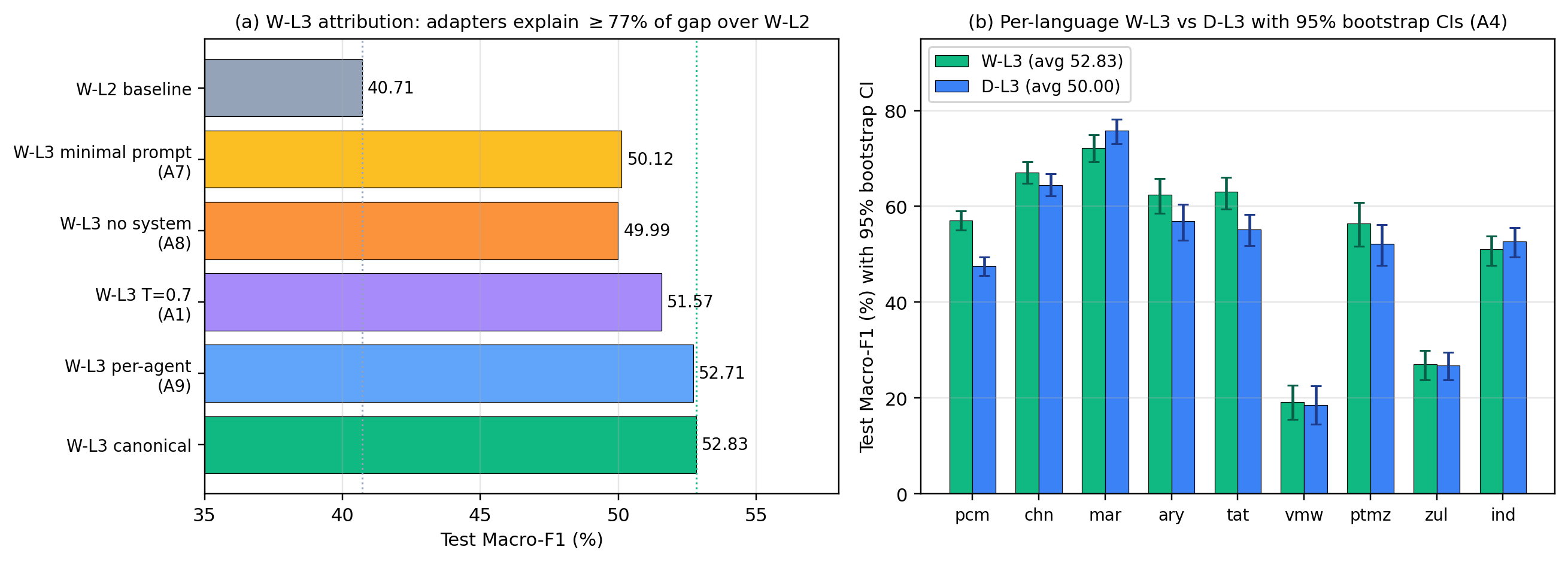}
  \caption{Ablation summary. Left: \wlthree remains strong under prompt and temperature ablations. Right: \wlthree vs. \dlthree per language with bootstrap confidence intervals.}
  \label{fig:ablation_appendix}
\end{figure*}

Figure~\ref{fig:depth_appendix} visualizes the depth-layer dynamics behind the learned-depth inversion discussed in Section~\ref{sec:results}. Figure~\ref{fig:perlang_appendix} provides a complementary per-language view, showing that the strongest \wlthree gains appear on the lower-resource end of the evaluation set. Figure~\ref{fig:ablation_appendix} gives the ablation summary moved from the main text.

\section{Aggregation and Per-Emotion Detail}
\label{app:aggregation_emotion}

Table~\ref{tab:agg_appendix} reports post-hoc aggregation variants for Width cells. The main paper uses majority vote as the canonical topology-consistent rule for the controlled matrix. The variants here are diagnostic rather than new matrix cells: they show that prompt-only Width contains useful signal that majority vote does not fully exploit, and that \wlthree is mildly aggregation-sensitive under a more permissive \texttt{any 3} rule. Table~\ref{tab:emotion_appendix} reports per-emotion F1 for canonical \wlthree, showing that the best matrix cell remains uneven across emotion labels.

\begin{table}[t]
\centering
\small
\begin{tabular}{lrrr}
\toprule
\textbf{Cell} & \textbf{maj. 2/3} & \textbf{conf. max} & \textbf{any 3} \\
\midrule
\wlone & 40.66 & \textbf{44.39} & 41.21 \\
\wltwo & 40.71 & \textbf{45.12} & 41.79 \\
\wlthree & 52.83 & 52.91 & \textbf{53.96} \\
\bottomrule
\end{tabular}
\caption{Aggregation-rule sensitivity for \w cells (test Macro-F1, \%).}
\label{tab:agg_appendix}
\end{table}

\begin{table}[t]
\centering
\small
\begin{tabular}{lr}
\toprule
\textbf{Emotion} & \textbf{\wlthree test F1} \\
\midrule
anger & 48.11 \\
disgust & 31.42 \\
fear & 49.75 \\
joy & 71.80 \\
sadness & 67.13 \\
surprise & 48.76 \\
\bottomrule
\end{tabular}
\caption{Per-emotion test F1 (\%) for \wlthree, averaged over languages.}
\label{tab:emotion_appendix}
\end{table}

\paragraph{Aggregation notation.} Let \(\hat{y}^{(k)}_i\in\{0,1\}^{|\mathcal{E}|}\) denote the binary emotion vector emitted by Width agent \(k\in\{1,2,3\}\) for example \(i\), and let \(c^{(k)}_{i,e}\in[0,1]\) denote the confidence score emitted by agent \(k\) for emotion \(e\). The canonical majority rule used in the main matrix is
\[
\hat{y}^{\mathrm{maj}}_{i,e}
=
\mathbf{1}\!\left[
\sum_{k=1}^{3}\hat{y}^{(k)}_{i,e}\ge 2
\right].
\]
The \texttt{any 3} rule emits a label when any agent emits it:
\[
\hat{y}^{\mathrm{any}}_{i,e}
=
\mathbf{1}\!\left[
\sum_{k=1}^{3}\hat{y}^{(k)}_{i,e}\ge 1
\right].
\]
The \texttt{confidence\_max} rule selects, for each label independently, the decision of the agent with the highest reported confidence:
\[
k^*_{i,e}=\arg\max_{k\in\{1,2,3\}} c^{(k)}_{i,e},
\qquad
\hat{y}^{\mathrm{conf}}_{i,e}
=
\hat{y}^{(k^*_{i,e})}_{i,e}.
\]

\section{Prompt and Reasoning Protocol}
\label{app:reasoning-protocol}

This section documents the prompt and reasoning protocol used by all matrix cells. The goal is to make the fixed components of the experiment auditable: the same reasoning directive, language-specific reasoning bank, decision policy, and JSON output schema are reused across Width and Depth, across all three diversity levels. This is important because the main paper attributes performance differences to topology and diversity source, not to hidden changes in task instructions or output format. The fixed-budget comparison applies to the six matrix cells: each uses exactly three LLM calls per sample. Baselines are reported with their own call counts, so the comparison to CoT+self-consistency at \(k=7\) is intentionally conservative with respect to call budget.

\subsection{Shared-call notation} For each example \(i\), let \(x_i\) be the input text, \(L_i\in\mathcal{L}\) the language identifier, \(b_{L_i}\) the rendered reasoning bank for that language, and \(y_i\in\{0,1\}^{|\mathcal{E}|}\) the gold multi-label vector over the six emotions. Every call receives the shared input package
\[
u_i=(x_i,L_i,b_{L_i}).
\]
Each model response is parsed as a JSON object
\[
z_i=(\pi(z_i),\kappa(z_i),r_i),
\]
where \(\pi(z_i)\in\{0,1\}^{|\mathcal{E}|}\) is the binary emotion vector, \(\kappa(z_i)\in[0,1]^{|\mathcal{E}|}\) is the confidence vector, and \(r_i\) is the short reasoning field. Only \(\pi(z_i)\) is used for scoring; \(\kappa(z_i)\) is used only in post-hoc aggregation analyses, and \(r_i\) is used for prompting and qualitative analysis.

\subsection{CoT block and decision policy}
\label{app:cot-block}

Table~\ref{tab:cot-block} gives the shared CoT block. The decision policy is intentionally conservative: it asks the model to identify evidence first, default to a single dominant emotion, and assign positive labels only when evidence is compelling. This policy is held fixed across all six matrix cells.

\begin{table}[t]
\centering\small
\begin{tabular}{p{0.92\columnwidth}}
\toprule
CoT block (verbatim, identical across all six cells) \\
\midrule
\texttt{Before classifying, reason step by step about the emotions in this text.} \\[2pt]
\texttt{Language-specific guidance for \{lang\}:} \\
\texttt{\{REASONING\_BANK\}} \\[2pt]
\texttt{Decision policy:} \\
\texttt{- Identify emotional cues first (words, phrases, emoji, tone)} \\
\texttt{- Consider cultural and linguistic context for \{lang\}} \\
\texttt{- Default to a single dominant emotion unless two distinct cue clusters clearly justify multiple labels} \\
\texttt{- Assign 1 only when evidence is compelling; when uncertain, assign 0} \\
\bottomrule
\end{tabular}
\caption{The CoT block injected as the system-prompt prefix in every Width and Depth call. \texttt{\{lang\}} is the BRIGHTER language code; \texttt{\{REASONING\_BANK\}} is the rendered per-language bank described in Appendix~\ref{app:reasoning-banks}.}
\label{tab:cot-block}
\end{table}

\subsection{Task instruction and JSON output schema}
\label{app:task-template}

Table~\ref{tab:task-instruction} gives the invariant task instruction and JSON schema. The schema emits one binary decision per emotion, one confidence score per emotion, and a short reasoning field. This lets us re-aggregate saved predictions under alternative aggregation rules without rerunning inference.

\begin{table}[t]
\centering\small
\begin{tabular}{p{0.92\columnwidth}}
\toprule
Task instruction (verbatim user-turn template) \\
\midrule
\texttt{Task: Classify the emotions expressed in the following text. The text may express zero, one, or multiple emotions simultaneously.} \\[2pt]
\texttt{Emotion categories: anger, disgust, fear, joy, sadness, surprise} \\[2pt]
\texttt{For each emotion, decide if it is present (1) or absent (0). Provide a confidence score between 0.0 and 1.0, and brief reasoning.} \\[2pt]
\texttt{Respond ONLY with a valid JSON object in exactly this format, with no additional text before or after:} \\
\texttt{\{} \\
\texttt{~~"emotions": \{"anger": 0, "disgust": 0, "fear": 0, "joy": 0, "sadness": 0, "surprise": 0\},} \\
\texttt{~~"confidence": \{"anger": 0.0, "disgust": 0.0, "fear": 0.0, "joy": 0.0, "sadness": 0.0, "surprise": 0.0\},} \\
\texttt{~~"reasoning": "Your analysis here."} \\
\texttt{\}} \\[2pt]
\texttt{Text: \{INPUT\_TEXT\}} \\
\bottomrule
\end{tabular}
\caption{The task instruction and JSON schema delivered through the user turn. This template is identical across all six matrix cells.}
\label{tab:task-instruction}
\end{table}

\section{Width Topology Prompt Designs}
\label{app:width-prompts}

Every Width cell composes its system prompt as an optional analytical lens followed by the shared CoT block in Table~\ref{tab:cot-block}. W-L1 and W-L3 use no analytical lens: their system prompt is only the shared CoT block. W-L2 uses three additive analytical lenses. The lenses are additive rather than restrictive: each agent is encouraged to attend to one evidence type, but is not forbidden from using the others.

\begin{table}[t]
\centering\small
\begin{tabular}{p{0.92\columnwidth}}
\toprule
W-L2 Agent A --- \emph{Surface-Priority} lens \\
\midrule
\texttt{You are an expert emotion detection system. When analyzing text, pay particular attention to explicit emotional signals: emotional keywords, sentiment-bearing phrases, emoji, emoticons, punctuation patterns (exclamation marks, repeated characters, ellipses), capitalization, and interjections. These surface cues should be your primary evidence, though you may also consider contextual meaning where relevant.} \\
\bottomrule
\end{tabular}
\caption{Surface-Priority lens system prompt for the W-L2 surface agent. The lens is concatenated with the language-specific CoT block in Table~\ref{tab:cot-block}.}
\label{tab:lens-surface}
\end{table}

\begin{table}[t]
\centering\small
\begin{tabular}{p{0.92\columnwidth}}
\toprule
W-L2 Agent B --- \emph{Context-Priority} lens \\
\midrule
\texttt{You are an expert emotion detection system. When analyzing text, pay particular attention to contextual and pragmatic meaning: implied emotions, overall tone, potential sarcasm or irony, rhetorical questions, cultural and idiomatic expressions, and what the author likely feels even if not stated explicitly. Look beyond literal word meanings, though you may also consider surface cues where relevant.} \\
\bottomrule
\end{tabular}
\caption{Context-Priority lens system prompt for the W-L2 context agent.}
\label{tab:lens-context}
\end{table}

\begin{table}[t]
\centering\small
\begin{tabular}{p{0.92\columnwidth}}
\toprule
W-L2 Agent C --- \emph{Structure-Priority} lens \\
\midrule
\texttt{You are an expert emotion detection system. When analyzing text, pay particular attention to how emotional signals interact across the text: sentiment shifts between clauses, contrasts or contradictions, co-occurring emotions that may reinforce or conflict with each other, and the overall emotional arc. Consider whether the text expresses one coherent emotion or multiple interacting emotions, though you may also consider individual cues where relevant.} \\
\bottomrule
\end{tabular}
\caption{Structure-Priority lens system prompt for the W-L2 structure agent.}
\label{tab:lens-structure}
\end{table}

W-L1 agents differ only through decoding stochasticity at \(T=0.7\). W-L3 agents share the same prompt form but load different QLoRA adapters at \(T=0.3\). Thus, the diversity in W-L3 is parametric rather than prompt-level.

\section{Depth Topology Inter-Layer Prompts}
\label{app:depth-prompts}

Depth cells emit three sequential calls. Layer~1 sees the input directly. Layer~2 receives the original input and Layer~1's serialized JSON output; Layer~3 receives the original input and the serialized JSON outputs from both previous layers. This makes the earlier prediction an external artifact that later calls can inspect, revise, or confirm. The depth topology also uses a descending temperature schedule, \(0.7 \rightarrow 0.5 \rightarrow 0.3\), implementing an exploration-to-consolidation pattern.

\subsection{Depth-chain notation} For diversity level \(\delta\in\{\mathrm{L1},\mathrm{L2},\mathrm{L3}\}\), Depth executes three JSON-conditioned calls:
\[
z_i^{(1)} = f^{\delta}_{1}(u_i;T_1),
\]
\[
z_i^{(2)} = f^{\delta}_{2}(u_i,z_i^{(1)};T_2),
\]
\[
z_i^{(3)} = f^{\delta}_{3}(u_i,z_i^{(1)},z_i^{(2)};T_3).
\]
The final Depth prediction is extracted from the third JSON output:
\[
\hat{y}^{D}_i=\pi(z_i^{(3)}).
\]
The fixed temperature schedule is
\[
(T_1,T_2,T_3)=(0.7,0.5,0.3),
\]
which implements the exploration-to-consolidation pattern described in the main paper.

\subsection{Inter-layer conditioning template}

Table~\ref{tab:depth-conditioning} gives the user-turn template used for inter-layer conditioning. For D-L1, all three layers share the same generic role and differ through sequential conditioning and temperature. For D-L2 and D-L3, Layers~2 and~3 additionally receive the Critic and Calibrator role prompts in Tables~\ref{tab:depth-critic} and~\ref{tab:depth-calibrator}.

\begin{table}[t]
\centering\small
\begin{tabular}{p{0.92\columnwidth}}
\toprule
Inter-layer conditioning user-turn template (Layers 2 and 3) \\
\midrule
\texttt{A previous analysis of the following text produced this result:} \\
\texttt{---} \\
\texttt{\{PRIOR\_JSON\}} \\
\texttt{---} \\[2pt]
\texttt{Now analyze the same text yourself. You may agree with, modify, or completely revise the previous analysis. Provide your own classification.} \\[2pt]
\texttt{\{TASK\_INSTRUCTION\}} \\[2pt]
\texttt{Text: \{INPUT\_TEXT\}} \\
\bottomrule
\end{tabular}
\caption{Inter-layer conditioning template prepended to the user message of Layers~2 and~3 in the Depth topology. \texttt{\{PRIOR\_JSON\}} contains Layer~1's full JSON output for Layer~2, and both earlier layers' full JSON outputs for Layer~3. The instruction permits agreement, modification, or full revision; the model is not forced to dissent.}
\label{tab:depth-conditioning}
\end{table}

\subsection{D-L2 and D-L3 role prompts}

\begin{table}[t]
\centering\small
\begin{tabular}{p{0.92\columnwidth}}
\toprule
Layer 2 Critic role (D-L2 and D-L3) \\
\midrule
\texttt{You are a critical reviewer of emotion analysis. Your job is to find and correct errors in a previous analysis. When reviewing, specifically check for:} \\
\texttt{(a) Emotions that may have been missed because they are expressed implicitly, culturally, or through understatement.} \\
\texttt{(b) Emotions that may have been incorrectly assigned due to surface-level pattern matching without considering context, sarcasm, or irony.} \\
\texttt{(c) Confidence scores that seem miscalibrated relative to the evidence.} \\
\texttt{(d) Inconsistencies between the reasoning and the predicted labels.} \\[2pt]
\texttt{If the previous analysis is correct, confirm it and explain why. Do not change predictions without justification.} \\
\bottomrule
\end{tabular}
\caption{Critic system prompt for Layer~2 in D-L2 and D-L3. The role prompt is concatenated with the shared CoT block in Table~\ref{tab:cot-block}.}
\label{tab:depth-critic}
\end{table}

\begin{table}[t]
\centering\small
\begin{tabular}{p{0.92\columnwidth}}
\toprule
Layer 3 Calibrator role (D-L2 and D-L3) \\
\midrule
\texttt{You are performing final calibration on an emotion analysis that has undergone two rounds of review. Your task is to:} \\
\texttt{(a) Confirm emotions where the evidence is strong and both prior analyses agree.} \\
\texttt{(b) Adjudicate where the two analyses disagree, weighing the evidence for each side.} \\
\texttt{(c) Remove emotions assigned on weak or ambiguous evidence --- prioritize precision over recall.} \\
\texttt{(d) Ensure confidence scores accurately reflect the strength of evidence.} \\
\bottomrule
\end{tabular}
\caption{Calibrator system prompt for Layer~3 in D-L2 and D-L3. The D-L3 truncation results in Table~\ref{tab:truncation} show that this final learned calibration stage can become harmful when the first learned specialist is already strong.}
\label{tab:depth-calibrator}
\end{table}

\section{Adapter Training Details}
\label{app:adapter_training}
\label{app:learned-adapters}

This appendix records the learned-specialization implementation. The detailed table below gives the Qwen2.5-14B-Instruct \citep{yang2024qwen25} adapter configuration. The second-backbone experiment instantiates the same Width specialist roles and Depth position-specific objectives on Llama-3.1-8B-Instruct \citep{grattafiori2024llama3} using the same pooled BRIGHTER training split; model-specific seed runs are reported in Appendix~\ref{app:seed_robustness}. The purpose of L3 is to make specialization parametric while keeping the number of inference calls and output protocol fixed within each backbone.

\begin{table*}[t]
\centering
\small
\resizebox{\textwidth}{!}{
\begin{tabular}{llp{3.0cm}p{3.5cm}p{3.0cm}p{3.2cm}}
\toprule
Cell & Adapter & Input conditioning & Training objective / weighting & QLoRA setup & Optimization setup \\
\midrule
W-L3 &
\texttt{general} &
Text + language reasoning bank &
Uniform supervised BCE on pooled Track-A training data &
$r=8$, $\alpha=16$, dropout $0.05$, 4-bit NF4 double quantization &
AdamW; LR $=10^{-5}$; batch $=4$; grad. accum. $=4$; effective batch $=16$; cosine schedule; warmup ratio $=0.10$; max epochs $=5$ \\

W-L3 &
\texttt{ambiguity} &
Text + language reasoning bank &
Supervised BCE; multi-label or hard examples weighted $\times 2$ &
$r=8$, $\alpha=16$, dropout $0.05$, 4-bit NF4 double quantization &
AdamW; LR $=10^{-5}$; batch $=4$; grad. accum. $=4$; effective batch $=16$; cosine schedule; warmup ratio $=0.10$; early-stopped at 3 epochs \\

W-L3 &
\texttt{contrastive} &
Text + language reasoning bank &
Supervised BCE plus cosine-hinge contrastive loss over emotion-present vs. emotion-absent representations &
$r=8$, $\alpha=16$, dropout $0.05$, 4-bit NF4 double quantization &
AdamW; LR $=10^{-5}$; batch $=4$; grad. accum. $=4$; effective batch $=16$; cosine schedule; warmup ratio $=0.10$; max epochs $=5$ \\

D-L3 &
\texttt{layer1} &
Text + language reasoning bank &
Uniform supervised BCE on pooled Track-A training data &
$r=8$, $\alpha=16$, dropout $0.05$, 4-bit NF4 double quantization &
AdamW; LR $=10^{-5}$; batch $=4$; grad. accum. $=4$; effective batch $=16$; cosine schedule; warmup ratio $=0.10$; max epochs $=5$; early-stopping patience $=2$ \\

D-L3 &
\texttt{layer2} &
Text + Layer-1 JSON output + language reasoning bank &
Supervised BCE; examples mispredicted by Layer 1 weighted $\times 3.0$ &
$r=8$, $\alpha=16$, dropout $0.05$, 4-bit NF4 double quantization &
AdamW; LR $=10^{-5}$; batch $=4$; grad. accum. $=4$; effective batch $=16$; cosine schedule; warmup ratio $=0.10$; max epochs $=5$; early-stopping patience $=2$ \\

D-L3 &
\texttt{layer3} &
Text + Layer-2 JSON output + language reasoning bank &
Supervised BCE; examples mispredicted by Layer 2 receive calibration-scaled additional weight $0.6$ &
$r=8$, $\alpha=16$, dropout $0.05$, 4-bit NF4 double quantization &
AdamW; LR $=10^{-5}$; batch $=4$; grad. accum. $=4$; effective batch $=16$; cosine schedule; warmup ratio $=0.10$; max epochs $=5$; early-stopping patience $=2$ \\
\bottomrule
\end{tabular}
}
\caption{QLoRA adapter training setup for the Qwen learned-specialization cells. All adapters use the same frozen Qwen2.5-14B-Instruct backbone, LoRA applied to \texttt{q\_proj}, \texttt{k\_proj}, \texttt{v\_proj}, \texttt{o\_proj}, \texttt{gate\_proj}, \texttt{up\_proj}, and \texttt{down\_proj}, bf16 compute, loss masking over non-assistant tokens, and the shared JSON output schema. W-L3 adapters use temperature $0.3$, top-$p=0.95$, and seed 42 at inference. D-L3 uses the layer-wise temperature schedule $(T_1,T_2,T_3)=(0.7,0.5,0.3)$, with top-$p=0.95$ and seed 42 at each layer. Zulu and Indonesian have no Track-C training rows and are not used in adapter training.}
\label{tab:adapter_hparams}
\end{table*}

\subsection{Qwen shared training recipe}

All Qwen L3 adapters use the same base recipe: QLoRA over the frozen Qwen2.5-14B-Instruct backbone, NF4 quantization with double quantization, LoRA rank \(r=8\), \(\alpha=16\), dropout \(0.05\), AdamW optimization, bf16 compute, and early stopping on development loss. The training data are the pooled public training rows over the seven languages with Track-A training data; Zulu and Indonesian have no train rows under the Track-C protocol. Loss is computed only over the assistant turn, with system and user tokens masked.

\paragraph{Adapter notation.} Let \(M_{\theta_0}\) denote the frozen Qwen2.5-14B-Instruct backbone. For Width-L3, let \(\mathcal{S}_{W}=\{\mathrm{gen},\mathrm{amb},\mathrm{ctr}\}\) denote the \texttt{general}, \texttt{ambiguity}, and \texttt{contrastive} adapters. For Depth-L3, let \(\phi_1,\phi_2,\phi_3\) denote the Layer~1, Layer~2, and Layer~3 adapter parameters.

\paragraph{Adapter parameterization.} For an adapted linear map with frozen weight \(W_0\), QLoRA uses the effective weight
\[
W_s = Q(W_0)+\Delta W_s,
\qquad
\Delta W_s=\frac{\alpha}{r}B_sA_s,
\]
where \(Q(W_0)\) is the NF4-quantized frozen weight, \(A_s\) and \(B_s\) are trainable low-rank matrices, \(r=8\), and \(\alpha=16\). Only the adapter parameters \(\phi_s=\{A_s,B_s\}\) are updated; the backbone parameters \(\theta_0\) remain frozen.

\paragraph{Assistant-token objective.} Each training example is serialized as a chat sequence \(c_i=[m_i^{\mathrm{sys}},m_i^{\mathrm{usr}},m_i^{\mathrm{asst}}]\), where the assistant turn contains the target JSON object. Let \(\mathcal{A}_i\) be the set of token positions belonging to the assistant turn. The weighted supervised loss for specialist \(s\) is
\[
\mathcal{L}_{\mathrm{sup}}^{(s)}
=
-\sum_{i=1}^{N}w_i^{(s)}
\sum_{t\in\mathcal{A}_i}
\log p_{\theta_0,\phi_s}(c_{i,t}\mid c_{i,<t}).
\]
System and user tokens are masked out of the loss, so the model is optimized only to generate the assistant-side JSON response.

\subsection{Width-L3 specialists}

The Width-L3 cell uses three parallel specialist adapters, abbreviated as \(\mathrm{gen}\), \(\mathrm{amb}\), and \(\mathrm{ctr}\), corresponding to \texttt{general}, \texttt{ambiguity}, and \texttt{contrastive}.

\noindent\textbf{\texttt{general}.} This adapter is trained uniformly on the pooled training set and acts as the baseline supervised specialist. Its objective is
\[
\mathcal{L}_{\mathrm{gen}}
=
\mathcal{L}_{\mathrm{sup}}^{(\mathrm{gen})},
\qquad
w_i^{(\mathrm{gen})}=1.
\]
It is the most important adapter in leave-one-out analysis: removing it reduces test Macro-F1 by 5.29 points.

\noindent\textbf{\texttt{ambiguity}.} This adapter upweights multi-label and hard examples, with sample weight doubled for multi-label or difficult cases. Let \(\mathcal{H}\) denote the set of hard examples identified for this specialist. Its example weight is
\[
w_i^{(\mathrm{amb})}
=
1+\mathbf{1}\!\left[\|y_i\|_1>1\ \lor\ i\in\mathcal{H}\right],
\]
so examples with multiple gold emotions or hard-case status receive weight \(2\). The objective is
\[
\mathcal{L}_{\mathrm{amb}}
=
\mathcal{L}_{\mathrm{sup}}^{(\mathrm{amb})}.
\]
It is intended to specialize in co-occurring emotions, where a single dominant-label prior may be insufficient. Removing it reduces test Macro-F1 by 2.01 points.

\noindent\textbf{\texttt{contrastive}.} This adapter is trained on the full data with an additional cosine-hinge contrastive term over emotion-present and emotion-absent representations. Let \(h_i\) denote the representation used for contrastive training, let \(P_i=\{e\in\mathcal{E}:y_{i,e}=1\}\) be the set of present emotions, and let \(N_i=\{e\in\mathcal{E}:y_{i,e}=0\}\) be the set of absent emotions. Let \(q_e\) denote the representation associated with emotion \(e\). The contrastive term is
\[
\begin{aligned}
\mathcal{L}_{\mathrm{ctr}}
&=
\sum_{i=1}^{N}
\sum_{e^+\in P_i}
\sum_{e^-\in N_i}
\ell_i(e^+,e^-),\\
\ell_i(e^+,e^-)
&=
\max\!\left(0,\,s_i(e^+,e^-)\right),\\
s_i(e^+,e^-)
&=
m-\cos(h_i,q_{e^+})+\cos(h_i,q_{e^-}).
\end{aligned}
\]
Here \(m\) is the hinge margin. The contrastive specialist minimizes
\[
\mathcal{L}_{\mathrm{ctr\text{-}adapter}}
=
\mathcal{L}_{\mathrm{sup}}^{(\mathrm{ctr})}
+
\lambda_{\mathrm{ctr}}\mathcal{L}_{\mathrm{ctr}}.
\]
It is intended to sharpen separation between affective categories and non-present labels. Removing it reduces test Macro-F1 by 4.78 points.

\subsection{Depth-L3 specialists}

The Depth-L3 cell uses three position-specific adapters. Layer~1 uses a uniform analyst adapter trained on the pooled public training set:
\[
\mathcal{L}_{1}
=
\mathcal{L}_{\mathrm{sup}}^{(1)}.
\]

Layer~2 uses a critic adapter trained to correct Layer~1 mistakes. Let
\[
\epsilon_i^{(1)}=\mathbf{1}\!\left[\pi(z_i^{(1)})\ne y_i\right]
\]
indicate whether Layer~1 makes an example-level prediction error. The Layer~2 training loader gives mistaken Layer~1 examples an effective weight of \(3.0\):
\[
w_i^{(2)}=1+2\epsilon_i^{(1)}.
\]
The Layer~2 objective is
\[
\begin{aligned}
\mathcal{L}_{2} ={}& -\sum_{i=1}^{N}w_i^{(2)}\sum_{t\in\mathcal{A}_i} \\
&\log p_{\theta_0,\phi_2}\!\left(c_{i,t}\mid c_{i,<t},z_i^{(1)}\right).
\end{aligned}
\]

Layer~3 uses a calibrator adapter trained on Layer~2 mistakes, but with a weaker correction weight. Let
\[
\epsilon_i^{(2)}=\mathbf{1}\!\left[\pi(z_i^{(2)})\ne y_i\right].
\]
The calibrator applies a calibration scaling of \(0.2\) to the Layer~2 error upweight, giving an additional error weight of \(0.6\):
\[
w_i^{(3)}=1+0.6\epsilon_i^{(2)}.
\]
Let \(z_i^{(<3)}=(z_i^{(1)},z_i^{(2)})\) denote the previous Depth outputs available to Layer~3. The Layer~3 objective is
\[
\begin{aligned}
\mathcal{L}_{3} ={}& -\sum_{i=1}^{N}w_i^{(3)}\sum_{t\in\mathcal{A}_i} \\
&\log p_{\theta_0,\phi_3}\!\left(c_{i,t}\mid c_{i,<t},z_i^{(<3)}\right).
\end{aligned}
\]

The truncation results in Table~\ref{tab:truncation} show that this learned correction chain is miscalibrated once the first specialist is already strong: D-L3 Layer~1 alone reaches 52.11 test Macro-F1, while the full learned chain reaches 50.00.

\subsection{Multi-LoRA serving}

All L3 inference uses a single vLLM engine with LoRA support enabled. The frozen base model is loaded once, and adapters are hot-swapped per call through LoRA requests. This means W-L3 preserves the same three-call inference budget as W-L1 and W-L2: it does not require loading or serving three separate full models.

For Width-L3, the three adapter predictions are
\[
\hat{y}^{(k)}_i
=
\pi\!\left(f_{\theta_0,\phi_k}(u_i)\right),
\qquad
k\in\mathcal{S}_{W}.
\]
The final prediction is the same per-label majority vote used by all Width cells:
\[
\hat{y}^{W\text{-}L3}_{i,e}
=
\mathbf{1}\!\left[
\sum_{k=1}^{3}\hat{y}^{(k)}_{i,e}\ge 2
\right].
\]

For Depth-L3, the position-specific adapters are applied sequentially:
\[
z_i^{(1)}
=
f_{\theta_0,\phi_1}(u_i),
\]
\[
z_i^{(2)}
=
f_{\theta_0,\phi_2}(u_i,z_i^{(1)}),
\]
\[
z_i^{(3)}
=
f_{\theta_0,\phi_3}(u_i,z_i^{(1)},z_i^{(2)}),
\]
with final prediction
\[
\hat{y}^{D\text{-}L3}_{i}
=
\pi(z_i^{(3)}).
\]

\section{Source Code and Configuration Map}
\label{app:src-map}

The source code, prompts, configurations, evaluation scripts, and experiment utilities are publicly available at \url{https://github.com/eracoding/topologyxdiversity}. Table~\ref{tab:src-map} maps the main experimental components to their implementation paths. The repository mirrors this structure, with prompts, configuration files, prediction outputs, and ablation scripts included for reproducibility.

\paragraph{Software environment.}
The reproducibility package includes a \texttt{requirements.txt} file specifying the minimum software versions used by the implementation. The requirements cover model serving, data processing, evaluation, configuration parsing, and utility scripts. QLoRA adapter training additionally depends on the standard Hugging Face/PEFT/bitsandbytes stack. The main dependencies are summarized in Table~\ref{tab:software_env}. Experiment-specific settings, including decoding parameters, adapter paths, aggregation rules, and evaluation scripts, are provided in the corresponding configuration files and source-code directories.

\begin{table}[t]
\centering
\small
\begin{tabular}{ll}
\hline
Component & Version requirement \\
\hline
vLLM serving & \texttt{vllm>=0.4.0} \\
PyTorch & \texttt{torch>=2.1.0} \\
Transformers & \texttt{transformers>=4.40.0} \\
PEFT & \texttt{peft>=0.10.0} \\
bitsandbytes & \texttt{bitsandbytes>=0.43.0} \\
Accelerate & \texttt{accelerate>=0.27.0} \\
pandas & \texttt{pandas>=2.0.0} \\
NumPy & \texttt{numpy>=1.24.0} \\
scikit-learn & \texttt{scikit-learn>=1.3.0} \\
PyYAML & \texttt{pyyaml>=6.0} \\
tqdm & \texttt{tqdm>=4.65.0} \\
jsonlines & \texttt{jsonlines>=4.0.0} \\
matplotlib & optional, \texttt{matplotlib>=3.7.0} \\
seaborn & optional, \texttt{seaborn>=0.12.0} \\
SciPy & optional, \texttt{scipy>=1.11.0} \\
\hline
\end{tabular}
\caption{Minimum software requirements used for serving, adapter training, data
processing, evaluation, and analysis. Optional visualization packages are not
required to reproduce the main numerical results.}
\label{tab:software_env}
\end{table}

\begin{table}[t]
\centering\small
\setlength{\tabcolsep}{3pt}
\begin{tabular}{p{0.34\columnwidth} p{0.58\columnwidth}}
\toprule
\textbf{Component} & \textbf{Source path} \\
\midrule
Shared base config & \path{config/base.yaml} \\
W-L1 / W-L2 / W-L3 configs & \path{config/config_b_l{1,2,3}.yaml} \\
D-L1 / D-L2 / D-L3 configs & \path{config/config_c_l{1,2,3}.yaml} \\
Width topology orchestrator & \path{src/topologies/width.py} \\
Depth topology orchestrator & \path{src/topologies/depth.py} \\
Base and prompt agents & \path{src/agents/base_agent.py}, \texttt{prompt\_agent.py} \\
LoRA agent & \path{src/agents/lora_agent.py} \\
W-L2 lens prompts & \path{src/prompts/system_prompts.py} \\
Depth role prompts & \path{src/prompts/inter_layer.py} \\
CoT block builder & \path{src/prompts/cot_enhancement.py} \\
Task instruction and JSON schema & \path{src/prompts/task_template.py} \\
Per-label majority vote & \path{src/aggregation/majority_vote.py} \\
Reasoning banks & \path{data/reasoning_banks/{lang}_*.json} \\
Shot banks & \path{data/shot_banks/{lang}_*.json} \\
Width-L3 adapters & \path{models/adapters/b_l3_{1,2,3}/} \\
Depth-L3 adapters & \path{models/adapters/c_l3_layer{1,2,3}/} \\
Adapter training scripts & \path{training/train_b_adapters.py}, \path{training/train_c_adapters.py} \\
Per-cell runners & \path{experiments/run_{b,c}_l{1,2,3}.py} \\
Results & \path{outputs/results/} \\
Ablation suite & \path{ablation/{a1,...,a13}/} \\
\bottomrule
\end{tabular}
\caption{Source-code and configuration map for the components referenced in the main paper.}
\label{tab:src-map}
\end{table}

\end{document}